\documentclass{article} % For LaTeX2e
\usepackage{intervene3d,times}

\usepackage{amsmath,amsfonts,bm}

\def\eqref#1{equation~\ref{#1}}
\def\1{\bm{1}}

\DeclareMathAlphabet{\mathsfit}{\encodingdefault}{\sfdefault}{m}{sl}
\SetMathAlphabet{\mathsfit}{bold}{\encodingdefault}{\sfdefault}{bx}{n}

\usepackage{hyperref}
\usepackage{url}
\usepackage{amsmath}
\usepackage{amssymb}
\usepackage{booktabs}
\usepackage{graphicx}
\usepackage{placeins}
\usepackage{amsthm}
\usepackage{xcolor}
\usepackage{colortbl}
\usepackage{tikz}
\usetikzlibrary{positioning,arrows.meta}
\usepackage[framemethod=default]{mdframed}
\definecolor{defbg}{RGB}{255,250,233}
\definecolor{defrule}{RGB}{191,143,24}
\definecolor{defhead}{RGB}{124,86,8}
\definecolor{thmbg}{RGB}{247,247,247}
\newtheoremstyle{defstyle}% name
  {0pt}{0pt}%                    space above and below, handled by the frame
  {\normalfont}%                 body font
  {}%                            indent
  {\bfseries\color{defhead}}%    head font
  {.}%                           punctuation after head
  {0.5em}%                       space after head
  {}%                            head spec
\mdfdefinestyle{defbox}{%
  backgroundcolor=defbg, linecolor=defrule, linewidth=3pt,
  topline=false, bottomline=false, rightline=false, leftline=true,
  innertopmargin=13pt, innerbottommargin=8pt,
  innerleftmargin=11pt, innerrightmargin=10pt,
  skipabove=9pt, skipbelow=9pt,
  nobreak=true}
\mdfdefinestyle{thmbox}{%
  backgroundcolor=thmbg, linecolor=black!45, linewidth=3pt,
  topline=false, bottomline=false, rightline=false, leftline=true,
  innertopmargin=13pt, innerbottommargin=8pt,
  innerleftmargin=11pt, innerrightmargin=10pt,
  skipabove=9pt, skipbelow=9pt,
  nobreak=true}
\theoremstyle{defstyle}
\newmdtheoremenv[style=defbox]{definition}{Definition}
\theoremstyle{plain}
\newmdtheoremenv[style=thmbox]{proposition}{Proposition}
\usepackage{wrapfig}
\usepackage{enumitem}
\usepackage{needspace}
\definecolor{oursink}{HTML}{2166AC}    % Intervene3D, proposed method
\definecolor{baseink}{HTML}{B2182B}    % predictor, main baseline
\definecolor{groupink}{HTML}{053061}   % navy: band titles
\definecolor{groupband}{HTML}{D1E5F0}  % band rows that group a table
\definecolor{oursband}{HTML}{EAF3F9}   % rows of Intervene3D
\newcommand{\oursrow}{\rowcolor{oursband}}              % start a row of Intervene3D
\newcommand{\oursname}[1]{\textcolor{oursink}{\textbf{#1}}}
\newcommand{\groupband}[2]{\rowcolor{groupband}\multicolumn{#1}{@{}l}{\textcolor{groupink}{\textbf{#2}}}\\}
\newcommand{\hdr}[1]{\begin{tabular}[b]{@{}c@{}}#1\end{tabular}}   % two-line column header

\title{Less Than Meets the Eye:\\Action-Set Identifiability and Principled\\Abstention for 3D Physical Geometry}

\author{Reshad Ul Karim\textsuperscript{1}, Abrar Samin\textsuperscript{1}, Sammam Mahdi\textsuperscript{1}, \\
\textbf{Syeda Maliha Tabassum\textsuperscript{1}, Md Rashad Al Hasan Rony\textsuperscript{2}} \\
\textsuperscript{1}BRAC University \qquad \textsuperscript{2}Deutsche Telekom AG \\
{\small\texttt{\{reshad.ul.karim, abrar.samin, sammam.mahdi\}@g.bracu.ac.bd}} \\
{\small\texttt{syeda.maliha.tabassum1@g.bracu.ac.bd, rah.rony@gmail.com}}}

\renewcommand{\textfraction}{0.12}
\renewcommand{\floatpagefraction}{0.75}

\prefinalcopy
\begin{document}

\maketitle
\suppressfloats[t]

\begin{abstract}
A corridor, its mirror reflection, and its image on a display can look identical, yet their physical surfaces lie at different depths. We test whether bounded camera motions
distinguish these explanations. Intervene3D takes an image and an initial geometry estimate,
predicts the view under each hypothesis and candidate motion, and selects a motion by maximin
separability. It reports the mechanism and first physical surface only if the available
motions separate the hypotheses still in contention. Otherwise it abstains. An adequacy test
also rejects observations that no hypothesis explains. On a matched synthetic benchmark with exact landmarks, disabling abstention gives confident answers on $76.4\%$ of unresolvable scenes. With abstention, none receives a confident answer under exact or noisy landmarks.
At $100\%$ accuracy on answered scenes, maximin selection answers $61.0\%$ of scenes and summed selection $43.1\%$. The adequacy test lowers wrong answers on strongly curved
mirror scenes from $56.8\%$ to $12.4\%$. Real display experiments compare two hypotheses
using one frozen depth-network estimate. The second image supplies photometric evidence and
is never passed through the depth network. One executed motion reduces the pooled median
surface error of twelve networks by $31.5\%$. These networks have larger error inside the
annotated panel than outside on $33.4\%$ of illusion-benchmark images. They place the first physical surface behind the glass on $36.9\%$ of Mirror3D frames.
\end{abstract}

\section{Introduction}
\label{sec:introduction}

\begin{figure}[!t]
\centering
\includegraphics[width=0.90\linewidth]{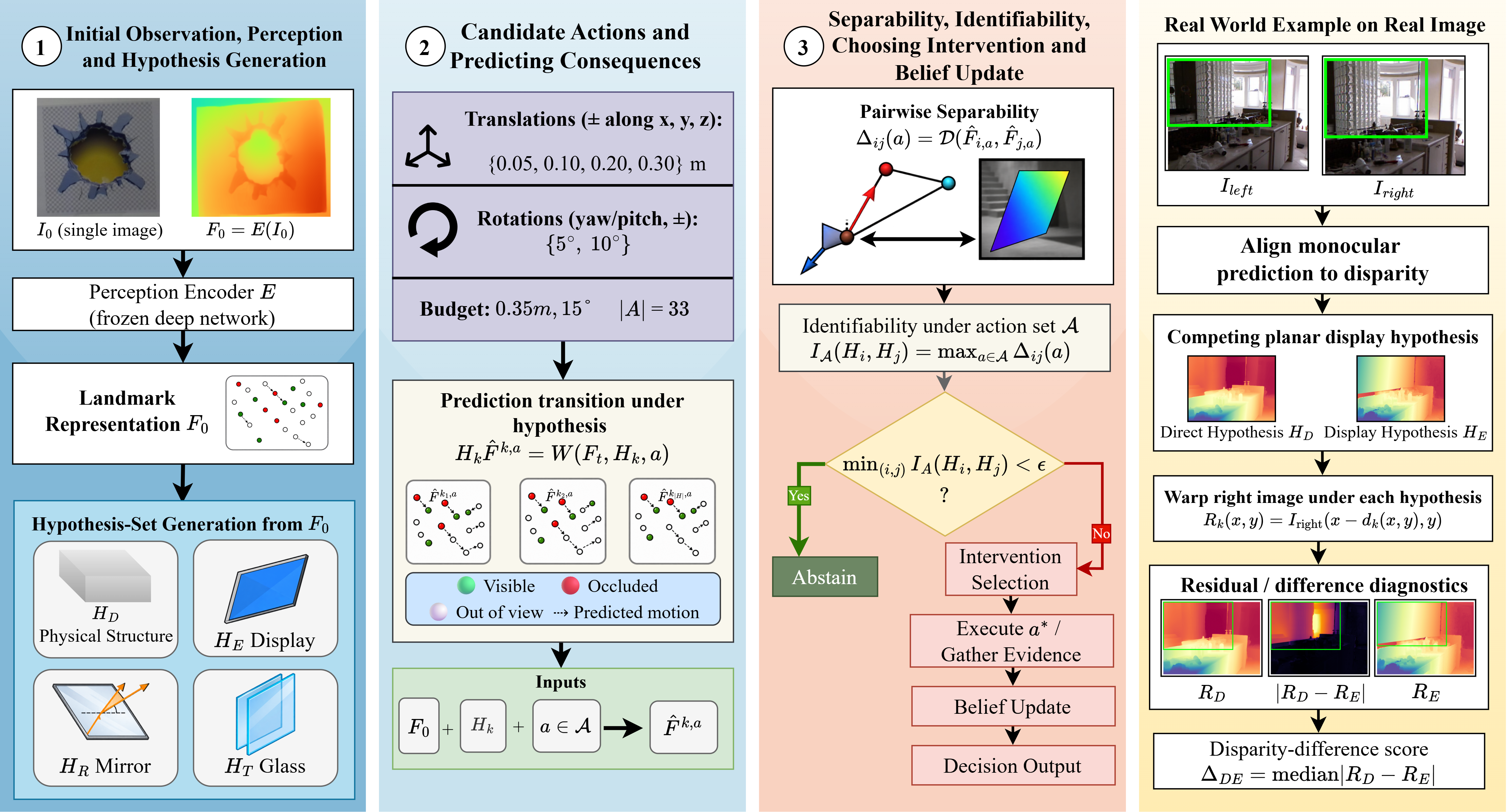}
\caption{\textbf{Intervene3D overview.} (1) The frozen-network prediction $F_0=E(I_0)$ of the initial image yields four hypotheses: direct content $H_D$, a flat emissive panel $H_E$, a planar mirror $H_R$, and a transparent pane $H_T$. (2) Each hypothesis predicts the observation after every candidate action in $\mathcal{A}$. (3) The method abstains when some contending pair is separated by no available action. Table~\ref{tab:protocol-order} states when each protocol applies this test. The right panel follows one real stereo pair: its right-camera image is warped under $H_D$ and $H_E$, and the photometric separability $\Delta_{DE}$ is the median difference of the two warps in gray levels. Green outlines mark the annotated region.}
\label{fig:system}
\end{figure}

A corridor viewed through an open doorway, reflected in a planar mirror, displayed on a
screen, or seen through a window can produce the same image. The first physical surface
along a camera ray is the nearest surface that would stop a physical probe: corridor structure,
mirror glass, the display panel, or the window pane. A depth network may estimate the
\emph{depicted} corridor behind a display even though the \emph{physical} surface is the panel.
When the images are identical, the initial image provides no evidence for choosing among
these mechanisms. Camera motion can expose their different light paths.
Matched hypotheses produce the same initial image and differ only after motion. Figure~\ref{fig:system} summarizes the method and shows, on one real stereo pair, the two hypothesis warps that are compared.
Several lines of work address mirrors, glass, multiple surfaces, and abstention. Multi-layer formulations retain both a transparent interface and the scene behind that interface \citep{layereddepth2025,seegroup2026,depthfocus2026}. Detectors localize mirrors and glass \citep{mirror3d2021,glint2026}. Selective prediction with split conformal prediction \citep{vovk2005,lei2018} decides when a system should answer. These methods estimate
surfaces and confidence from a fixed viewpoint. This work asks whether any available viewpoint can identify the physical mechanism. Identifiability depends on the hypothesis family, the
bounded action set, and a separability threshold with units specified by the protocol.

\paragraph{Input and output.}Intervene3D takes one image, its geometry estimate, and candidate
camera motions. Real data also supplies an annotated region, camera calibration, and stereo
reference disparity outside that region for alignment. A frozen monocular depth network runs
on the initial image only. Each hypothesis predicts how image content moves. Warped photographs provide the evidence after motion. The output is a mechanism and the first
physical surface, or an abstention.
The full family contains direct content ($H_D$), a flat emissive panel ($H_E$), a planar
mirror ($H_R$), and a transparent pane ($H_T$), all matched at the initial pose. The basic
transmission model gives $H_T$ and $H_D$ identical image motion, so motion alone cannot
distinguish them. The real illusion benchmark \citep{illusion2025} contains printed sheets,
posters, and monitors in photographed scenes. Experiments on it use only $H_D$ and $H_E$. Two selectors choose the motion. Intervene3D-maximin selects the motion that maximizes the smallest separability among separable, contending hypothesis pairs. Intervene3D-sum uses their summed separability. With two hypotheses, maximin and summed selection coincide. Evidence that the selectors differ therefore comes from the synthetic benchmark.
After the selected motion, residuals determine the answer. The system also checks whether
the action set can separate the remaining alternatives, and applies an adequacy test where
enabled. Table~\ref{tab:protocol-order} states when each protocol performs these checks.

\begin{enumerate}[nosep,leftmargin=*,topsep=2pt]

\item \textbf{Action-set identifiability, intervention selection, and an adequacy test.} We
define separability under camera motion and test whether every contending hypothesis pair
can be separated by an available action. A maximin selector chooses the motion. An adequacy test with a threshold fixed in advance flags observations the family does not explain.

\item \textbf{A benchmark with measured ambiguity.} The three base mechanisms are
pixel-identical at the initial view to below $3\times10^{-13}$~px. Scenes that no action resolves allow abstention to be scored. Variants add a dielectric pane and mechanisms outside
the base family.
\item \textbf{Real camera motion.} Registered frames of $53$ real scenes supply executed camera motions for $135$ regions in $44$ of these scenes. A calibrated stereo baseline supplies a fixed viewpoint change on $404$ photographs. A hand-held RGB-D capture adds a monitor and a mirror.
\item \textbf{Measurements across networks and scenes.} Errors on depicted content are measured across twelve checkpoints, and physical mirror surfaces are evaluated on $123$ Mirror3D frames. One illusion category uses a rebuilt stereo reference, validated under a fixed
acceptance rule.

\end{enumerate}

\section{Method}
\label{sec:method}

\begin{figure}[t]
\centering
\begin{tikzpicture}
\node[anchor=south west, inner sep=0] (img) at (0,0) {\includegraphics[width=0.95\linewidth]{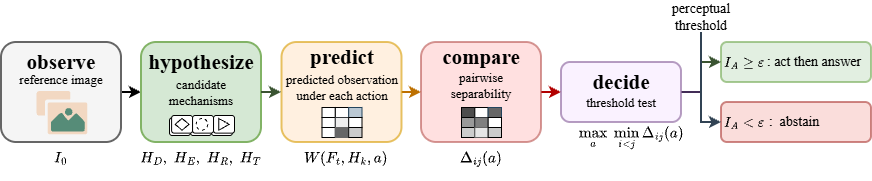}};
\begin{scope}[x={(img.south east)}, y={(img.north west)}]
\fill[white] (0.652,0.09) rectangle (0.797,0.302);
\node[font=\scriptsize] at (0.722,0.185) {$\min_{i<j}\max_{a}\Delta_{ij}(a)$};
\end{scope}
\end{tikzpicture}
\caption{\textbf{Decision pipeline.} The initial image $I_0$ yields the four hypotheses, and each hypothesis predicts the observation after every action $a$. Separability $\Delta_{ij}(a)$ compares those predictions, and the smallest identifiability over contending pairs is compared with $\epsilon$. Table~\ref{tab:protocol-order} states when each protocol applies this test. The synthetic protocol adds an adequacy check ($\eta=2$~px) and a confidence check on the updated belief $p_1$ after the executed motion.}
\label{fig:pipeline}
\end{figure}

Figure~\ref{fig:pipeline} shows the decision pipeline. The real-data protocols
instantiate separability and residuals through image warps (Section~\ref{sec:real-protocol}).

\subsection{Hypotheses and the Action Set}
\label{sec:formulation}

Let $I_0$ be the initial image at camera pose $C_0$, and $F_0=E(I_0)$ its geometry estimate.
On real images, $E$ is a frozen monocular depth network and $F_0$ becomes a dense disparity
map after alignment. On the synthetic benchmark, $F_0$ consists of exact or noisy landmarks with image coordinates, depths, and visibility flags (Section~\ref{sec:setup}).
\begin{definition}[Image-formation hypothesis family]\label{def:hypotheses}
Intervene3D considers $\mathcal H=\{H_D,H_E,H_R,H_T\}$: direct content, a flat emissive panel,
a planar mirror, and a transparent pane. Each hypothesis specifies a first physical surface.
The hypotheses are \emph{matched}: they give the same image at $C_0$. Differences after motion therefore arise from the hypothesis. The basic $H_T$ and $H_D$ predict identical image motion. Separating them requires an additional cue, such as the pane reflection in
Section~\ref{sec:res-family}.
\end{definition}
The panel, mirror, and pane interfaces are planar. Direct content and content behind an
interface retain the full three-dimensional structure in $F_0$. An action $a\in SE(3)$
applies a rigid camera motion from $C_0$. The bounded set $\mathcal A$ contains translations,
rotations, and the null action (Section~\ref{sec:setup}). All actions are evaluated by enumeration. Identifiability depends on $(\mathcal H,\mathcal A,\epsilon)$ through \eqref{eq:identifiability}.

\subsection{Predicting the Consequence of an Action}
\label{sec:transition}

The transition $W$ maps the initial geometry, a hypothesis, and a candidate action to predicted geometry,
\begin{equation}
\label{eq:transition}
\hat{F}^{\,k,a} \;=\; W\!\left(F_0,\, H_k,\, a\right),
\end{equation}
where $H_k$ is the $k$-th hypothesis and $\hat F^{k,a}$ contains its projected landmarks and
visibility after $a$. Equation~\ref{eq:transition} uses one shared projection for every hypothesis. Only the hypothesized world changes. A display uses a plane-induced homography, and a planar mirror uses exact reflection
across its plane. In the paraxial limit, a plane-parallel pane forms a fixed apparent image, which yields the motion equivalence in Definition~\ref{def:hypotheses}. Closed-form transitions
keep the initial views matched to numerical precision. A curved surface with known parameters
can also retain an exact transition (Section~\ref{sec:res-family}). Learned transitions can introduce differences between fitted models.

\subsection{Separability, Identifiability, and Abstention}
\label{sec:separability}

Equation~\ref{eq:separability} defines pairwise separability as the distance between the geometries that hypotheses $H_i$ and $H_j$ predict for the same action. On the synthetic benchmark,
\begin{align}
\label{eq:separability}
\Delta_{ij}(a) \;&=\; D\!\left(\hat{F}^{\,i,a},\, \hat{F}^{\,j,a}\right), \\
\label{eq:distance}
D \;&=\; \lambda_m D_{\text{motion}} + \lambda_o D_{\text{occlusion}} .
\end{align}
The motion term is the root-mean-square displacement of jointly visible landmarks in pixels.
The occlusion term counts landmarks with different visibility under the two hypotheses. Visibility depends on image bounds and the finite interface aperture. The weights give $D$
pixel-equivalent units.
A hypothesis pair is separable under $a$ when $\Delta_{ij}(a)\geq\epsilon$, where $\epsilon$
is the separability threshold. The contending pairs $\mathcal C$ are the hypothesis pairs that still require a decision. Table~\ref{tab:protocol-order} defines this set for each protocol. Identifiability $\mathcal I_{\mathcal A}$ is a pair's best separability over the action set. The system abstains if some
contending pair cannot be separated by any action:
\begin{align}
\label{eq:identifiability}
\mathcal{I}_{\mathcal{A}}\!\left(H_i, H_j\right)
\;&=\; \max_{a \in \mathcal{A}} \; \Delta_{ij}(a), \\
\label{eq:abstain}
\min_{(i,j) \in \mathcal{C}} \; \mathcal{I}_{\mathcal{A}}\!\left(H_i, H_j\right) \;&<\; \epsilon .
\end{align}
The minimum over pairs of the maximum over actions reaches $\epsilon$ only when every
contending pair has a separating action, possibly a different action for each pair.
Separability uses pixel-equivalent landmark distance in simulation, median pixel displacement
under real executed motions, and median gray-level difference for stereo. Each protocol's
$\epsilon$ has the corresponding unit. These scores are not numerically interchangeable.

\begin{table}[t]
\caption{\textbf{Contention and decision order by protocol.} The synthetic confidence and adequacy checks are specified in Section~\ref{sec:belief}; real protocols choose the smaller photometric residual.}
\label{tab:protocol-order}
\centering
\footnotesize\setlength{\tabcolsep}{3pt}
\begin{tabular}{@{}p{2.1cm}p{5.0cm}p{5.9cm}@{}}
\toprule
Protocol & Contending hypothesis pairs $\mathcal C$ & When \eqref{eq:abstain} is applied \\
\midrule
Synthetic & $p_i p_j>10^{-3}$ after the belief update; if none remain, the maximum-posterior hypothesis against each alternative & Select and execute one motion, update the belief, then test the remaining alternatives \\
Executed real motion & Median relative first-surface depth difference exceeds $2\%$, determined before motion & Test predicted separability over registered candidate poses before using the selected target frame \\
Fixed stereo & The two hypotheses $H_D,H_E$ on the annotated region & Test photometric separability using the already captured right image \\
\bottomrule
\end{tabular}
\end{table}

\subsection{Choosing the Intervention}
\label{sec:selection}

Two objectives are compared. Intervene3D-sum maximizes summed separability,
\begin{equation}
\label{eq:sum}
a^{\star}_{\text{sum}} \;=\; \arg\max_{a \in \mathcal{A}} \;\sum_{i<j} p_i\, p_j\,
\Delta_{ij}(a),
\end{equation}
where $p_i,p_j$ are the current hypothesis probabilities. This sum can favor an action that
leaves the weakest contending pair unseparated. Equation~\ref{eq:maximin} defines Intervene3D-maximin through a minimum over pairs,
\begin{equation}
\label{eq:maximin}
a^{\star}_{\text{maximin}} \;=\; \arg\max_{a \in \mathcal{A}} \;\min_{(i,j) \in \mathcal{C}_{\epsilon}} \;
\Delta_{ij}(a) / w_{ij},
\end{equation}
where $w_{ij}=p_i p_j/\max_{(r,s)\in\mathcal C_\epsilon}(p_r p_s)$ normalizes joint mass. Selection excludes pairs
with $p_i p_j\leq10^{-3}$. The set $\mathcal C_\epsilon$ further keeps only pairs with
$\mathcal I_{\mathcal A}(H_i,H_j)\geq\epsilon$. Otherwise, an inseparable pair could force the minimum to zero for every action. The abstention rule \eqref{eq:abstain} handles that ambiguity. If no
separable pair remains, selection falls back to \eqref{eq:sum}.
All reported one-motion experiments select from a uniform prior, so $w_{ij}=1$ at selection.
The comparison is therefore maximin versus summed separability. The maximin model-discrimination criterion is related to $T$-optimality in optimal design. It can differ from the sum with
three or more hypotheses: three pairs in the three-mechanism benchmark, six in the
four-mechanism benchmark. With two hypotheses the objectives select the same action.

\subsection{Belief Update, Output, and the Adequacy Test}
\label{sec:belief}

A residual compares one hypothesis with the image $O'$ captured after the selected motion. On the synthetic benchmark, the simulator supplies observed landmarks $F_{\mathrm{obs}}$. The residual $e_k=D(\hat F^{k,a^\star},F_{\mathrm{obs}})$ uses the same landmark distance as separability. On real data, residuals compare warped photographs with the observed image (Section~\ref{sec:real-protocol}). They use no second depth-network estimate.
The synthetic belief update is
\begin{equation}
\label{eq:belief}
p_1(H_k) \;\propto\; \exp\!\left(-\beta\, e_k\right)\, p_0(H_k),
\end{equation}
where $\beta$ converts the pixel-equivalent residual into a log-likelihood and $p_0$ is the prior.
A posterior floor of $10^{-6}$ prevents one noisy measurement from eliminating a hypothesis.
The synthetic system abstains if \eqref{eq:abstain} holds, if $\max_k p_1(H_k)\leq\tau$ for the commitment threshold $\tau$,
or if the two largest posteriors tie (numerical tolerance $10^{-9}$). When the adequacy test is enabled, the system also abstains if $\min_k e_k>\eta$, where the adequacy threshold $\eta=2$~px is fixed before evaluation. The adequacy test can flag an out-of-family mechanism only as unexplained. If a family member reproduces it within the adequacy threshold under every action, the observations cannot distinguish the two. The test is evaluated separately from the main synthetic results and is not applied in the
reported real-data protocols. A passing decision reports the most probable mechanism and
its first physical surface: direct structure for $H_D$, the panel for $H_E$, mirror glass for
$H_R$, or the pane for $H_T$.

\paragraph{Separability and residual.} Separability compares two hypotheses under one motion. A large residual for $H_D$ shows disagreement between the initial depth estimate and the new image.
A replacement surface is returned when a different hypothesis wins and the decision checks pass.

\subsection{Protocols for Real Data}
\label{sec:real-protocol}

Both illusion-benchmark protocols use $\mathcal H=\{H_D,H_E\}$. The calibrated stereo baseline supplies one fixed viewpoint change, $|\mathcal A|=1$. Registered photographs of a static scene supply a larger action set. Executed-motion separability is predicted before motion. Stereo separability uses the captured second photograph. The depth network runs once on the initial image. Each hypothesis uses that estimate to warp photographs, and the observed images provide the decision evidence. An image is admissible only when at least $90\%$ of the stereo reference disparity inside the region lies within $1$~px of a plane fitted by random sample consensus. The criterion checks the flat-panel assumption independently of the network estimate. Alignment first expresses the network output in inverse-depth form $\hat p$, converting metric-depth outputs as required. Scale $s$ and shift $t$ in Equation~\ref{eq:align} are fitted by least squares to stereo reference
disparity outside the annotated region, then applied inside it:
\begin{equation}
\label{eq:align}
d_{D} \;=\; s\,\hat{p} + t .
\end{equation}
Fit and evaluation regions are disjoint, so alignment cannot absorb the error being measured. The same stereo reference aligns both the reported surface and the depth estimate
used for comparison. For the flat-panel hypothesis $H_E$, the \emph{ring plane} is fitted to the aligned estimate on a $24$-pixel ring outside the region and evaluated inside it.
Fitting inside would inherit the depicted relief. The ring locates panels flush with their
surroundings, but can misplace a panel standing in front of the background
(Section~\ref{sec:res-stereo}).

\paragraph{Fixed stereo baseline.}In a rectified stereo pair, column $x$ in the left image
with disparity $d$ corresponds to $x-d$ in the right image. Equation~\ref{eq:warp} warps the right photograph into the left frame under each hypothesis $H_k$ with disparity $d_k$, and Equation~\ref{eq:delta-photometric} defines the photometric separability:
\begin{align}
\label{eq:warp}
R_{k}(x, y) \;&=\; I_{\text{right}}\!\left(x - d_{k}(x,y),\; y\right),
\qquad k \in \{D, E\}, \\
\label{eq:delta-photometric}
\Delta_{DE} \;&=\; \operatorname{median}\left|R_{D} - R_{E}\right| .
\end{align}
The median uses pixels valid in both warps.
The score compares image evidence: geometric disagreement on a featureless patch need not
produce a photometric difference. Both warps share the same source photograph, which reduces differences due to capture conditions. Sampling at different pixels can still introduce noise.
The separability $\Delta_{DE}$ is in gray levels, with $\epsilon=2.0$. Below this threshold the system abstains.
Otherwise it returns the hypothesis with smaller
$e_k=\operatorname{median}|R_k-I_{\text{left}}|$, also in gray levels.

\paragraph{Executed motions.}Every ordered frame pair of a static scene is registered by perspective-n-point with random sample consensus. A target frame is accepted with at least
$100$ inliers and median reprojection error at most $1$~px. Its pose relative to the initial
frame supplies one candidate motion. Each hypothesis predicts the movement of textured
region pixels. Separability is the median difference between the predicted displacements, with $\epsilon=1.0$~px.
Here $H_D,H_E$ contend when the median of $|z_D-z_E|/[(z_D+z_E)/2]$ exceeds $2\%$,
where $z_D,z_E$ are first-surface depths. If the surfaces agree within this tolerance, the
output is marked surface-equivalent. The lower-residual hypothesis then supplies the surface, and no mechanism is claimed. Otherwise \eqref{eq:abstain} is checked
before using the selected target frame. Photometric residuals compare the initial frame with
the target warped into it under each hypothesis. A shared brightness gain and offset are
fitted outside the region. The smaller residual determines the answer. These residuals are in gray levels, distinct
from the pixel units of separability. All thresholds are fixed before the reported runs.

\paragraph{Hand-held capture.} A hand-held RGB-D camera (Intel RealSense D435) records a photograph shown on a monitor inside a flat surround, and a wall mirror. Three printed markers give the camera pose in every frame. The hypothesis set is $\{H_D,H_E,H_R\}$. Separability, residuals and the decision follow the executed-motion protocol. A small light fixed on the camera serves as the observer cue. A hypothesis whose predicted reflection of the light contradicts the executed frame receives a residual penalty of $16$ gray levels. The light position is calibrated from the mirror recording, and each half of that recording is evaluated with the position fitted on the other half.

\section{Results}
\label{sec:results}

A matched synthetic benchmark evaluates motion selection, abstention, and mechanisms outside the
family. Real data then evaluate the decision and the reported surface with twelve released depth
networks, under executed camera motion, in a hand-held RGB-D capture, and on stereo photographs.
Figure~\ref{fig:main} summarizes failure rates (a), synthetic selection (b), and
executed-motion surface errors (c).

\subsection{Experimental Setup}
\label{sec:setup}

Synthetic variants of a base scene share the same initial rays and apparent depth. Their initial views coincide to within $2.3\times10^{-13}$~px, leaving a single-view classifier at chance. Marker points on the operator's rig provide a possible mirror cue after motion. They are resampled until their reflection lies outside the initial view. A scene is labeled resolvable when some action separates its true hypothesis from each alternative, with different actions allowed. Some scenes remain unresolvable within the action set.
Classifier splits follow base scenes.
Causal explanation accuracy (CEA) is the share of scenes with the correct mechanism,
counting abstention as an error. Committed accuracy is that share among answered scenes. The false physical certainty rate (FPCR) is the share of unresolvable scenes answered with posterior above $\tau$. Identifiability AUROC measures how well the identifiability score ranks resolvable against unresolvable scenes. Overall abstention and abstention on resolvable scenes are also reported. Synthetic values are means and sample standard deviations over $11$ seeds, with the benchmark redrawn per seed. Surface error is the median relative error of the reported first physical surface against the stereo reference inside the annotated region. It uses disparity for stereo photographs and depth for executed motions. An answer of $H_D$ retains the network estimate. A network is \emph{fooled} on an image when its median relative error inside the annotated region exceeds its error outside. This operational label does not identify a mechanism by itself. A region--checkpoint combination is one annotated region evaluated with one network checkpoint.
The action set includes translations of $\{0.05,0.10,0.20,0.30\}$~m along each of three axes
in both directions, yaw and pitch rotations of $\{5^\circ,10^\circ\}$ in both directions,
and the null action: $24+8+1=33$ actions. Equation~\ref{eq:distance} uses $\lambda_m=1$. Priors are uniform. The constants $\epsilon=1.0$~px, $\beta=1.0$, $\tau=0.45$, and $\lambda_o=4.0$~px are not tuned. Exact landmarks define the exact-encoder condition. The noisy
encoder adds $0.6$~px image-coordinate noise and $2\%$ relative depth noise.
On real data, twelve published checkpoints are evaluated as released \citep{depthanything2024,da32026,depthpro2025,moge2025,moge32026,metric3d2023,unidepth2024,unidepthv22025}. They span seven architecture families and a $38\times$ range in parameter count. None is trained or tuned on any evaluation set. A transparency-fine-tuned Depth Anything variant returns metric depth that requires a depth prompt. Without that prompt its alignment fails, so the variant is excluded from the twelve-checkpoint evaluation.

\subsection{Synthetic Benchmark}
\label{sec:res-synth}

\begin{figure}[t]
\centering
\includegraphics[width=\linewidth]{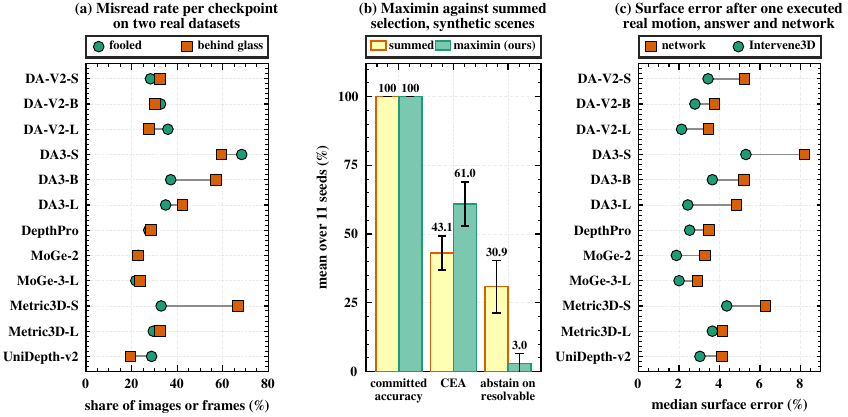}
\caption{\textbf{Main results.} (a) Fooled share on $404$ admissible illusion images per checkpoint and behind-glass share on $123$ Mirror3D frames (means $33.4\%$ and $36.9\%$). (b) Maximin and summed selection on the three-mechanism benchmark, mean and sample standard deviation over $11$ seeds. CEA equals answered share here because committed accuracy is $1.000$. (c) Median surface error after one executed real motion, $135$ regions per checkpoint (pooled medians $4.35\%$ for the network and $2.98\%$ for Intervene3D). Errors are percentages. The monitor-in-the-frame category uses rebuilt reference disparity.}
\label{fig:main}
\end{figure}

Table~\ref{tab:phase1} reports the three-mechanism benchmark ($H_D,H_E,H_R$).
The single-view classifier stays at chance in every seed. With exact landmarks, every
method that executes a motion and permits abstention answers correctly whenever it commits.
Figure~\ref{fig:main}b shows that maximin answers $61.0\%$ of scenes and summed selection $43.1\%$. The relative gain is $42\%$ at equal committed accuracy. On resolvable scenes, maximin abstains
on $3.0\%$, compared with $30.9\%$ for summed selection, $36.7\%$ for largest-translation,
and $67.2\%$ for random selection. Most remaining abstentions are on unresolvable scenes. On direct scenes, which contain no display or mirror, maximin answers $154$ of the $164$ resolvable scenes, and all $154$ answers are correct.
Maximin without abstention reaches CEA $0.799\pm0.045$, compared with $0.622\pm0.030$ for passive multi-view. This ordering holds in all $11$ seeds.
Disabling abstention for Intervene3D-sum gives confident answers on $76.4\%$ of unresolvable scenes. Enabling it reduces that share to $0.0\%$. The noisy encoder retains committed accuracy
$0.996$ and FPCR $0.000$. Execution perturbations reduce the CEA of Intervene3D-sum by $5.0\%$.

Abstention decreases as the action set grows. Widening the lateral budget from $0.30$~m to
$1.00$~m raises the resolvable share from $62.8\%$ to $72.8\%$ and lowers maximin abstention
from $39.0\%$ to $29.3\%$. Abstention on resolvable scenes stays near $3\%$, with committed
accuracy $1.000$. In the four-mechanism benchmark, $H_T$ and
$H_D$ have identical image motion. Maximin abstains on all direct and transmission scenes. It commits correctly on displays and mirrors.

\subsection{Beyond the Hypothesis Family}
\label{sec:res-family}

The first of two tests adds three primary out-of-family mechanisms: a spherical mirror, a cylindrical display,
and a water-like surface, plus a strongly curved mirror variant.
Without the adequacy test, maximin assigns $60$--$66\%$ of the three primary mechanisms to a family member. None of those answers is $H_D$. The $2$~px adequacy test lowers the share of wrongly answered scenes from $56.8\%$ to $12.4\%$ for strongly curved mirrors. Over the three primary mechanisms, it lowers that share from $62.5\%$ to $49.1\%$. It leaves in-family results unchanged
with exact landmarks. For the noisy encoder, it changes CEA to $0.385$, committed accuracy
to $1.000$, and abstention to $62\%$. Water-like surfaces are detected on only $4.2\%$ of
scenes because a family member, most often $H_R$, fits within $2$~px on about $96\%$.
A fifth hypothesis, a spherical mirror of known sag (curvature depth), answers curved-mirror scenes at accuracy $1.000$. A nominal sag gives wrong answers on $43\%$ of those scenes.
The second test models glass as a dielectric interface that transmits the scene and weakly
reflects the operator, with normal-incidence Fresnel reflectance of $5.4$--$13.8\%$.
Some action then separates the direct hypothesis from its alternatives on $48.1\%$ of direct
scenes, and the transmission hypothesis on $22.7\%$ of transmission scenes. Neither class is
resolvable under the basic transmission model. Committed accuracy remains $1.000$ across all
four mechanisms. Mirrors become harder to resolve because the pane now also reflects the
operator: their resolvable share falls from $69.3\%$ to $25.2\%$. Varied marker contrast
provides a cue because a pane hides dim reflected markers that a mirror shows. The transmission gain vanishes with constant marker contrast.

\subsection{Executed Camera Motion in Real Scenes}
\label{sec:res-motion}

The executed-motion experiment uses illusion-benchmark scenes photographed from several
positions. Of $906$ ordered frame pairs over $53$ scenes, $528$ pass the fixed registration
criteria, with median reprojection error $0.80$~px. Each of $135$ annotated initial regions
retains at least one accepted target frame. Selected motions have median rotation
$22.2^\circ$ and translation $0.204$~m. Table~\ref{tab:motion} reports the results.
\begin{table}[t]
\caption{\textbf{Executed real motions on the illusion benchmark.} Twelve checkpoints, $135$ regions each. The ground-truth mechanism is $H_E$ throughout, and the surface error compares the answered surface with the stereo reference. A region contends when $H_D$ and $H_E$ place the surface more than $2\%$ apart. The envelope keeps targets within $10^\circ$ and $0.30$~m of the initial frame, the synthetic action range.}
\label{tab:motion}
\centering
\footnotesize \setlength{\tabcolsep}{4pt}
\begin{tabular}{@{}lcc@{}}
\toprule
& Every accepted motion & Within the envelope \\
\midrule
Region--checkpoint combinations & $1{,}620$ & $732$ \\
Abstention & $0.0\%$ & $0.8\%$ \\
Median surface error, \textcolor{oursink}{answer} / \textcolor{baseink}{network} & $\mathbf{2.98\%}$ / $4.35\%$ & --- \\
Answer worse than the network & $9.3\%$ of combinations & --- \\
Checkpoints with lower median error & $\mathbf{12}$ of $12$ & --- \\
Mechanism accuracy, contending regions & $64.4\%$ of $663$ & --- \\
\bottomrule
\end{tabular}
\end{table}
Figure~\ref{fig:main}c shows that one executed motion lowers the pooled median surface error from $4.35\%$ to $2.98\%$. The relative reduction is $31.5\%$, and the median is lower on all twelve checkpoints.
The answer is worse than the network on $9.3\%$ of the $1{,}620$ region--checkpoint combinations. No combination abstains with every accepted motion. Abstention is $0.8\%$ within the $10^\circ$/$0.30$~m envelope
of the synthetic action set (Table~\ref{tab:motion}). Among the $663$ contending combinations, mechanism accuracy is $64.4\%$. Every wrong answer is $H_D$
on a physical panel. A post hoc analysis finds lower accuracy with larger pose reprojection
error, consistent with registration limiting the decision.
Selection has little effect here because maximin and summed selection coincide and almost every accepted motion separates the two hypotheses. Five selection rules differ by at most $1.4$ percentage points.
These data evaluate the decision and reported surface. Testing the advantage of maximin selection requires three or more contending hypotheses. The hand-held mirror recording below supplies such a case.

\begin{figure}[t]
\centering
\includegraphics[width=.95\linewidth]{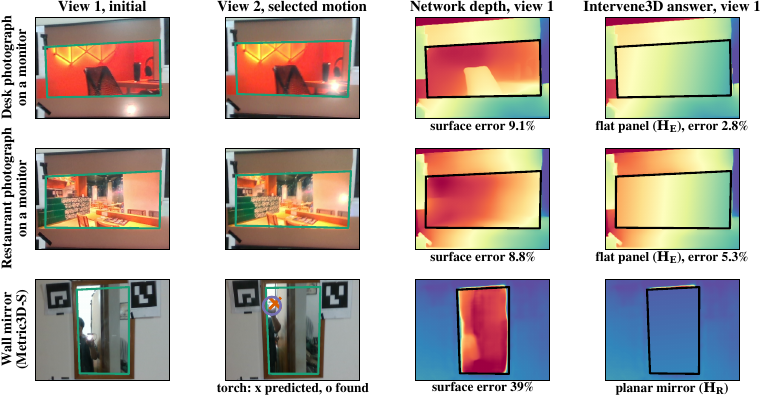}\par\vspace{3pt}
\includegraphics[width=\linewidth]{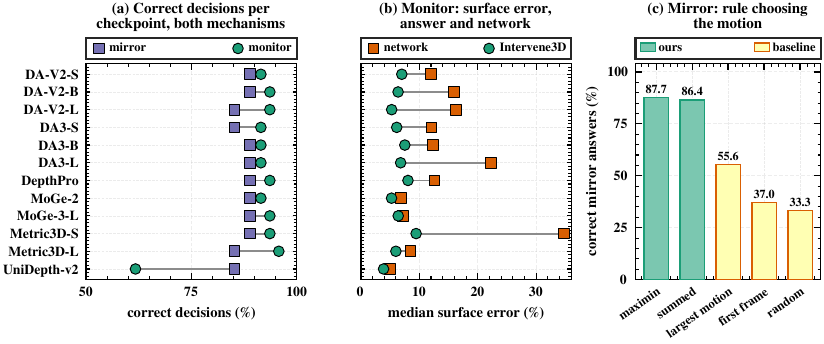}
\caption{\textbf{Hand-held capture.} Top: the median correctly answered region of each recording, with DA-V2-Small on the monitor. The reported surface replaces the network depth inside the region by the plane fitted on a surrounding ring. Bottom: (a) correct decisions over $564$ monitor and $324$ mirror combinations, (b) monitor surface error of the answers $H_E$, (c) mirror accuracy under five motion rules.}
\label{fig:c1-qual}
\label{fig:c1-quant}
\end{figure}

\paragraph{Hand-held capture.}Figure~\ref{fig:c1-qual} follows one
region of each recording through the decision. On the monitor, the network depth reproduces the depicted room, and the
reported surface is the flat panel. Figure~\ref{fig:c1-quant}b shows that answers of $H_E$ lower the pooled median
surface error from $11.5\%$ to $6.3\%$, with a lower error on all twelve checkpoints. Figure~\ref{fig:c1-quant}a shows
that the decision is correct on $90.2\%$ of the $564$ monitor region--checkpoint combinations ($47$ regions) and on $87.7\%$ of the $324$ mirror combinations ($27$ regions). Figure~\ref{fig:c1-quant}c shows that the choice of motion matters on the mirror: maximin
selection answers correctly on $87.7\%$, summed selection on $86.4\%$, the largest motion on $55.6\%$, the first frame
on $37.0\%$, and a random motion on $33.3\%$.

\subsection{Real Stereo Photographs}
\label{sec:res-stereo}

A calibrated stereo rig provides one fixed viewpoint change: a $63$~mm baseline along the
image rows. Table~\ref{tab:category} covers twelve checkpoints on $404$ admissible photographs,
giving $4{,}848$ image--checkpoint combinations. In the monitor-in-the-frame category, which
shows a video frame on a screen, the released reference disagrees with its stereo images by
about $3$~px inside the screen. This offset is constant over time and also affects static
bezel features. That category's reference is rebuilt from each stereo pair under a fixed acceptance rule. With that correction, the networks are
fooled on $33.4\%$ of images overall. Figure~\ref{fig:main}a gives the fooled share of each checkpoint.
\begin{table}[t]
\caption{\textbf{Illusion-benchmark stereo results} \citep{illusion2025}\textbf{.} Twelve checkpoints, $\epsilon=2.0$ gray levels. Shares pooled over checkpoints, with the checkpoint range. $H_E$ counts as correct because every region is a physical panel. Surface errors are medians. $^\dagger$Rebuilt reference. $^\ddagger$The ring plane lies behind the screen (Section~\ref{sec:res-stereo}).}
\label{tab:category}
\centering
\footnotesize \setlength{\tabcolsep}{1.2pt}
\begin{tabular}{@{}lrrrrcc@{}}
\toprule
& & & & & \multicolumn{2}{c}{Surface error (\%), \textcolor{oursink}{answer} / \textcolor{baseink}{network}} \\
\cmidrule(l){6-7}
Category & Images & \hdr{Fooled, \%\\(range)} & \hdr{Abstain,\\\%} & \hdr{Committed\\acc., \%} & \hdr{all\\answers} & \hdr{answers\\$H_E$} \\
\midrule
Printed sheet on a table & $88$ & $47.2$ ($30.7$--$75.0$) & $74.1$ & $64.1$ & $4.8$ / $8.2$ & $\mathbf{4.2}$ / $9.4$ \\
Poster beside an object & $119$ & $59.1$ ($47.9$--$73.9$) & $82.0$ & $67.3$ & $7.7$ / $13.4$ & $\mathbf{7.6}$ / $16.3$ \\
Monitor in the frame$^\dagger$ & $98$ & $11.7$ ($1.0$--$44.9$) & $87.8$ & $72.9$ & $1.2$ / $3.2$ & $\mathbf{1.1}$ / $4.1$ \\
Monitor filling the frame$^\ddagger$ & $99$ & $11.5$ ($0$--$97$) & $33.3$ & $2.5$ & $7.6$ / $7.5$ & --- \\
\midrule
\textbf{All admissible} & $\mathbf{404}$ & $\mathbf{33.4}$ ($21.8$--$68.3$) & $69.8$ & $32.3$ & --- & --- \\
\bottomrule
\end{tabular}
\end{table}
One stereo baseline separates the two hypotheses on a minority of images. The system abstains on the rest. Table~\ref{tab:category} shows that answers of $H_E$ have lower median surface error than the depth estimate in the first three categories. Including answers of $H_D$, the median error remains below the network's for all twelve checkpoints in each of those categories.
The fourth category exposes the ring fit's limitation. A monitor filling the frame stands in
front of the wall, so the ring plane lies behind the screen. The system mostly answers $H_D$,
which is counted as the wrong mechanism because the region is a panel. Its committed accuracy
is therefore only $2.5\%$, although the median surface error of $7.6\%$ is close to the
network's $7.5\%$.

\paragraph{Real mirrors and ordinal depth.}Mirror3D \citep{mirror3d2021} supplies $123$
frames with human-annotated mirror planes. The twelve checkpoints place the first physical
surface behind the glass on $36.9\%$ of frames on average, and are fooled on $69.1\%$. These frames evaluate depth networks only. They supply no camera motion for the decision protocol. LayeredDepth's externally authored ordinal layer labels provide
another network-only evaluation, with lower accuracy on multi-layer point pairs.

\subsection{What the Hypotheses Add}
\label{sec:res-adds}

Mirror and glass detectors, cross-view consistency checks, and learned failure classifiers flag suspect regions or depth estimates. Here each method is scored on the surface that it reports. When every region detected by an open-vocabulary detector \citep{owlv2} is replaced by the ring plane, the median surface error over the stereo region--checkpoint combinations is $2.84\%$. This replacement makes the network surface worse on $52.7\%$ of the combinations. When the network surface is kept wherever Intervene3D abstains, the median is $2.89\%$ and the surface is made worse on $0.8\%$. Under one executed motion, the detector reaches $3.00\%$ and worsens $25.8\%$ of the $1{,}620$ combinations, against $2.98\%$ and $9.3\%$ for Intervene3D. Within a category, cross-view disagreement predicts failure at AUROC $0.56$--$0.62$. A gradient-boosted failure classifier reaches AUROC $0.891$ with checkpoints held out and $0.859$ with checkpoints and scenes held out. When the regions it flags are replaced by the ring plane for a held-out category, the surface is made worse on $4.8\%$ of combinations, against $0.6\%$ for Intervene3D. Detector outputs can also supply the prior $p_0$ of Equation~\ref{eq:belief}.
The image-formation hypotheses additionally provide a candidate replacement surface and a criterion for whether the available views can distinguish them.

\section{Related Work}
\label{sec:related}

Prior work localizes mirrors and glass as scene elements
\citep{mirrornet2019,glassdetection2020,mirror3d2021}, models transparency through
decomposed radiance \citep{glint2026}, and quantifies the difficulty with dedicated
benchmarks \citep{booster2023,clearpose2022,dreds2022}. Robustness studies carry the same
concern to corrupted inputs \citep{robodepth2023}, and a recent challenge evaluates current systems on high-resolution depth for these materials \citep{ntire2025depth}.
Multi-layer formulations represent more than one surface along a ray
\citep{layereddepth2025,seegroup2026,depthfocus2026}. An illusion benchmark documents at
scale that depth estimates can follow depicted content \citep{illusion2025}.
Detector and multi-layer outputs can supply priors over the hypothesis family. This work tests whether the available observations determine the mechanism behind an appearance.
Classical analyses establish that image evidence alone can leave a family of scenes
indistinguishable \citep{kutulakos2008theory}. Active and next-best-view perception choose
viewpoints using belief-weighted objectives \citep{nbvreflective2022,rainforth2024}.
Intervene3D asks whether any view within a bounded set permits a mechanism decision. It interprets identifiability interventionally \citep{citris2022}.
Deferral rules \citep{geifman2017,selectivenet2019} and split conformal prediction
\citep{vovk2005,lei2018} control which predictions are retained. The failure classifiers in Section~\ref{sec:res-adds} estimate when depth is unreliable. Intervene3D
reports a mechanism and first physical surface when its decision checks pass, or abstains.

\section{Conclusion}
\label{sec:conclusion}

Intervene3D combines action-set identifiability, motion selection, and explicit abstention
to report the first physical surface. On matched synthetic scenes, maximin selection answers
more scenes than summed selection at equal committed accuracy. The adequacy test rejects
some mechanisms outside the family. A known-sag mirror hypothesis permits correct
answers on curved mirrors. Real two-hypothesis experiments show that image evidence after
motion can improve the reported surface across twelve frozen depth networks.

\paragraph{Limitations.}The decision depends on the specified family, action set, and threshold.
Most water-like surfaces in Section~\ref{sec:res-family} pass the adequacy test despite being
outside the family. Real motions are recovered from previously captured frames. The illusion
experiments have two hypotheses. Mirrors in video add a third but lack a cue separating the
mirror from an opening. Scenes without panels or mirrors are evaluated through the direct scenes of the synthetic benchmark. Stereo alignment uses reference disparity outside the annotated region. The ring fit can misplace a panel in front of its surroundings.

\bibliography{intervene3d_arxiv}
\bibliographystyle{intervene3d}

\clearpage
\clearpage
\appendix
\setcounter{topnumber}{2}
\setcounter{bottomnumber}{1}
\setcounter{totalnumber}{2}
\renewcommand{\floatpagefraction}{0.92}
\renewcommand{\textfraction}{0.08}
\section*{Supplementary Material}

\makeatletter
\let\supp@addcontentsline\addcontentsline
\def\supp@toc{toc}
\renewcommand\addcontentsline[3]{%
  \begingroup
    \def\supp@arg{#1}%
    \ifx\supp@arg\supp@toc
      \endgroup\supp@addcontentsline{stoc}{#2}{#3}%
    \else
      \endgroup\supp@addcontentsline{#1}{#2}{#3}%
    \fi}
\begingroup
  \renewcommand{\contentsname}{\normalsize Contents}
  \setlength{\parskip}{0pt}
  \setcounter{tocdepth}{2}
  \footnotesize
  \section*{\contentsname}
  \@starttoc{stoc}
\endgroup
\makeatother
\bigskip

\section{Experimental Details}
This appendix gives the notation, the reason one image admits several mechanisms, the simulation and real-photograph protocols, the metrics, the comparison methods, and the minimum resolving baseline.

\subsection{Notation}
\label{sec:supp-notation}

Table~\ref{tab:notation} collects the symbols of Section~\ref{sec:method}.

\begin{table}[htb]
\caption{\textbf{Method notation.} Separability and residual units depend on the protocol (Sections~\ref{sec:separability} and~\ref{sec:real-protocol}).}
\label{tab:notation}
\centering
\footnotesize \setlength{\tabcolsep}{3pt}
\begin{tabular}{@{}p{2.9cm}p{10.4cm}@{}}
\toprule
Symbol & Meaning \\
\midrule
$I_0,C_0$ & Initial image and camera pose \\
$E,F_0$ & Geometry source and initial estimate; a frozen depth network on real images, supplied landmarks in simulation \\
$H_D,H_E,H_R,H_T$ & Direct content; flat emissive panel; planar mirror; transparent pane \\
$\mathcal H$ & Hypothesis family; the real illusion protocols use $\{H_D,H_E\}$ \\
$\mathcal A,a$ & Bounded set of candidate camera motions and one motion \\
$W,\hat F^{k,a}$ & Closed-form transition and geometry predicted under $H_k$ after $a$ \\
$D,\lambda_m,\lambda_o$ & Synthetic landmark distance and its motion and occlusion weights \\
$\lambda_g,\lambda_f$ & Unused geometry and feature weights in the general distance \\
$\Delta_{ij}(a),\epsilon$ & Pairwise separability and its threshold in the same units \\
$\mathcal I_{\mathcal A}(H_i,H_j)$ & Largest separability of $H_i,H_j$ over the action set \\
$\mathcal C,\mathcal C_\epsilon$ & Contending pairs (Table~\ref{tab:protocol-order}); their separable subset \\
$w_{ij}$ & Joint hypothesis mass normalized by the largest joint mass; $1$ at selection in reported runs \\
$O',F_{\mathrm{obs}},e_k$ & New image; observed synthetic landmarks; residual for $H_k$ \\
$p_0,p_1,\beta,\tau$ & Prior, updated belief, inverse temperature, and commitment threshold \\
$\eta$ & Adequacy threshold, $2$~px in the synthetic adequacy experiments \\
$s,t,\hat p$ & Alignment scale, shift, and inverse-depth estimate \\
$d_D,d_E$ & Aligned direct disparity and fitted flat-panel disparity \\
$R_D,R_E,\Delta_{DE}$ & Warped photographs and their median absolute gray-level difference \\
\bottomrule
\end{tabular}
\end{table}

\subsection{Why One Image Admits Several Mechanisms}
\label{sec:supp-concept}

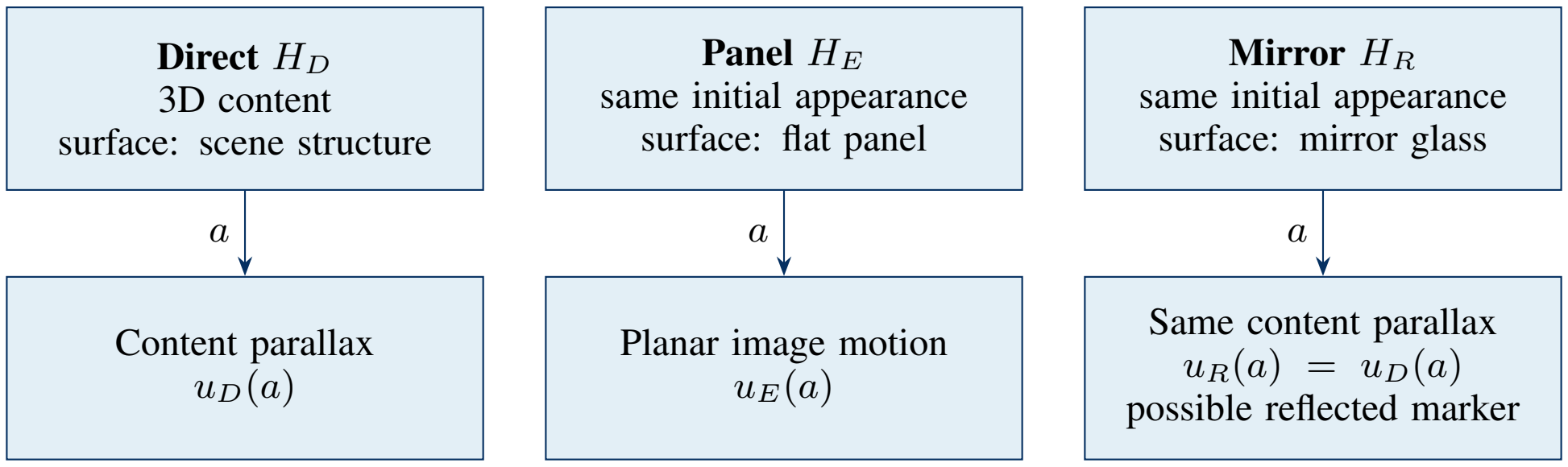
\begin{figure}[htb]
\centering
\begin{tikzpicture}[
  font=\small,
  box/.style={draw=groupink,fill=groupband!60,align=center,text width=37mm,minimum height=15mm,inner sep=3pt},
  arr/.style={-{Stealth[length=1.6mm]},draw=groupink},node distance=5mm]
\node[box] (direct) {\textbf{Direct $H_D$}\\3D content\\surface: scene structure};
\node[box,right=of direct] (display) {\textbf{Panel $H_E$}\\same initial appearance\\surface: flat panel};
\node[box,right=of display] (mirror) {\textbf{Mirror $H_R$}\\same initial appearance\\surface: mirror glass};
\node[box,below=7mm of direct] (directafter) {Content parallax\\$u_D(a)$};
\node[box,below=7mm of display] (displayafter) {Planar image motion\\$u_E(a)$};
\node[box,below=7mm of mirror] (mirrorafter) {Same content parallax\\$u_R(a)=u_D(a)$\\possible reflected marker};
\draw[arr] (direct) -- node[left]{$a$} (directafter);
\draw[arr] (display) -- node[left]{$a$} (displayafter);
\draw[arr] (mirror) -- node[left]{$a$} (mirrorafter);
\end{tikzpicture}
\caption{\textbf{Matched appearance and motion cues.} The three hypotheses share the initial image. Motion can distinguish a static panel from direct 3D content. Mirror content and direct content have equal parallax; distinguishing them requires a reflected operator cue. $u_k(a)$ denotes image motion under $H_k$.}
\label{fig:concept}
\end{figure}

Figure~\ref{fig:concept} distinguishes image motion from the first physical surface.
Direct content can span several depths. A display places the same appearance on one plane,
so its motion follows a plane-induced homography. A planar mirror produces the same content
parallax as an opening of the same shape. Only an additional cue, such as the operator's
reflection appearing after motion, separates those two hypotheses. Let $u_k(a)$ denote
content motion under $H_k$; it is distinct from pairwise separability $\Delta_{ij}(a)$.
The basic pane hypothesis has the same motion as direct content. A dielectric pane can add
weak reflected markers (Section~\ref{sec:res-family}).

\subsection{Simulation Setup}
\label{sec:supp-sim}

Table~\ref{tab:supp-sim} lists every parameter of the synthetic generator, read from the
configuration files of the reported runs. Each base scene draws one interface, one camera and
one operator rig. Every mechanism's variant of that scene is built from the same reference
rays and the same apparent depth, which makes the variants pixel-identical at $C_0$ by
construction. Scenes vary in interface distance, size, aspect and tilt, in the depth and depth
spread of the content behind the interface, in camera placement and pointing, and in where the
operator's reflection would appear. The content spread is the main difficulty control. The minimum resolving baseline of Proposition~\ref{prop:mcrb} scales with $1/|1/Z_1 - 1/Z_2|$,
so a shallow scene needs a large motion, and a sufficiently shallow one is unresolvable within
the action bounds. One display in four is view-tracked, and no motion resolves such a display within its aperture. The operator rig is offset so that its reflection lands near the aperture boundary. As a result, some mirrors are resolvable only by a large motion and others by none.

\begin{table}[htb]
\caption{\textbf{Parameters of the synthetic generator.} $U[a,b]$ denotes a uniform draw per base scene.}
\label{tab:supp-sim}
\centering
\footnotesize \setlength{\tabcolsep}{4pt}
\begin{tabular}{@{}lp{8.2cm}@{}}
\toprule
Parameter & Value \\
\midrule
Base scenes, seeds & $96$ per seed with three mechanisms, $192$ with four; $11$ seeds, benchmark redrawn per seed; splits by base scene, $50/16.7/33.3\%$ \\
Camera & $320 \times 240$~px, $f_x = f_y = 320$~px; interface center offset $U[-0.30, 0.30]$~m laterally and vertically relative to the camera, pointing jitter $U[-6^\circ, 6^\circ]$ \\
Interface & distance $U[1.2, 2.5]$~m; half-width $U[0.45, 1.15]$ of the image half-width at that distance; aspect $U[0.60, 0.95]$; tilt up to $8^\circ$ \\
Content & $40$ landmarks, jittered stratified sampling over $85\%$ of the interface extent in a random subset of a $7 \times 7$ grid; nearest depth behind the interface $U[0.02, 0.80]$~m, depth spread $U[0.02, 2.50]$~m \\
Display & static, or view-tracked with probability $0.25$ \\
Operator rig & $3$ markers; rig offset $U[-1.8, 1.8]$~m laterally and $U[-0.8, 0.8]$~m vertically, jitter $0.12$~m, forward offset $U[-0.10, 0.02]$~m; resampled until the reflection is out of view at $C_0$ \\
Pane (four mechanisms) & thickness $U[0.004, 0.060]$~m, refractive index $U[1.40, 1.75]$ \\
Dielectric pane & unpolarized two-interface Fresnel reflectance at the viewing angle, $5.4$ to $13.8\%$ at normal incidence; marker contrast $U[0.05, 1.0]$; a reflected marker is visible when reflectance times contrast is at least $0.02$ \\
Held-out mechanisms & sag over half-diagonal or half-width, log-uniform in $[0.005, 0.5]$, capped for the primary spherical mirror so that the operator's reflection stays out of frame at $C_0$; water-like surface with refractive index $1.333$ and half of the landmarks from the reflected scene \\
Actions & $33$: translations $\{0.05,0.10,0.20,0.30\}$~m on three axes in both directions, yaw and pitch rotations $\{5^\circ,10^\circ\}$ in both directions, and null; budgets of $49$ and $61$ actions in Appendix~\ref{sec:supp-budget} \\
\bottomrule
\end{tabular}
\end{table}

\subsection{The Real-Photograph Protocol in Detail}
\label{sec:supp-protocol}

\begin{figure}[htb]
\centering
\includegraphics[width=\linewidth]{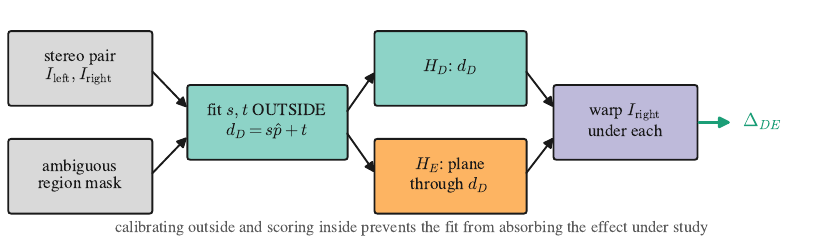}
\caption{\textbf{Construction of the stereo measurement.} The reference disparity is admitted when a plane fitted by random sample consensus holds at least $90\%$ of the region within $1$~px. Scale and shift of the prediction are fitted outside the region. $H_D$ is the aligned prediction and $H_E$ a plane fitted to it on a $24$~px ring outside the region. The right image is warped under each hypothesis, and $\Delta_{DE}$ is the median difference between the two warps.}
\label{fig:supp-protocol}
\end{figure}

Figure~\ref{fig:supp-protocol} presents the construction of the stereo measurement and the
place where each quantity is fitted. Hypotheses are compared through warped photographs:
depth disagreement can occur even on pixels whose appearance cannot distinguish the hypotheses.
A featureless patch can therefore have large depth disagreement but zero photometric
separability. The median uses pixels valid in both warps. Sharing a source photograph reduces capture differences. Noise at the different sampled locations need not cancel.

\paragraph{Admissibility.} A physical sheet or panel is flat, and the reference disparity inside
the region should lie on a plane. The criterion counts the share of region pixels within $1$~px
of a plane fitted by random sample consensus and admits the image at $90\%$. An independent
plane fit admits the same $405$ of $455$ images for every checkpoint; $404$ remain after the
monitor-in-the-frame reference is rebuilt. Before that rebuild, the fit admits $88$ of
$104$ sheets, $119$ of $153$ posters, and all $99$ images of each monitor category. An
ordinary least-squares fit scored over all pixels would instead drop every image of a monitor
filling the frame, because of a few boundary pixels.

\paragraph{Ring fit.} The ring of $H_E$ keeps the valid pixels within $24$~px outside the region. When fewer than $500$ remain, the fit uses every valid pixel outside the region.

\paragraph{The monitor-in-the-frame reference.} Inside the screen of every image of this
category, the released reference places the panel about $2.9$~px of disparity nearer than its
own stereo pair does. The offset has an interquartile range of $-3.01$ to $-2.77$~px and forms a smooth tilted plane identical across all six videos. Static bezel features carry the same offset as the moving
content, which rules out capture timing, and the reference map is also displaced by about
$35$ to $40$~px vertically against the image. Since the offset alone exceeds the typical error of
a predictor outside the region, a predictor that saw the screen exactly as the stereo pair does
would typically be labeled fooled. The reference is therefore rebuilt from each stereo pair with a full-resolution normalized cross-correlation matcher with a left-right check, whose parameters are frozen before use. The rebuild is validated on the monitor-filling-the-frame category, whose released reference agrees with its stereo pair. It is accepted only if the mean over images of the share of rebuilt region pixels within $1$~px of that reference reaches $0.85$. The statistic reaches $0.872$. The rebuilt reference is sparse, valid on $30\%$ of the region pixels of the validation category, and never interpolated. Every error inside the region uses valid pixels alone. Under it, $98$ of $99$ images remain admissible.

\subsection{Metric Definitions}
\label{sec:supp-metrics}

\begin{table}[htb]
\caption{\textbf{The metrics reported throughout the paper.} Here $\hat{H}$ denotes the predicted
mechanism, $H^\star$ the true one, and $y_{\mathrm{id}} = 0$ a case that no action resolves.}
\label{tab:supp-metrics}
\centering
\footnotesize
\setlength{\tabcolsep}{3pt}
\begin{tabular}{@{}lp{4.4cm}p{5.0cm}@{}}
\toprule
Metric & Definition & Question answered \\
\midrule
CEA & $P(\hat{H} = H^\star)$ & how often the mechanism is named correctly, abstention counted as an error \\
Mechanism accuracy & $P(\hat H=H^\star)$ over the stated evaluation subset & mechanism correctness over all regions or the contending subset, as specified \\
Committed acc. & $P(\hat{H} = H^\star \mid \text{answered})$ & how often a given answer is correct \\
Abstention rate & $P(\text{abstain})$ & how often an answer is withheld \\
Abst.\ on resolvable & $P(\text{abstain} \mid y_{\mathrm{id}} = 1)$ & how often an answer is withheld that some action would warrant \\
FPCR & $P(\max_k p(H_k) > \tau,\ \text{answered} \mid y_{\mathrm{id}} = 0)$ & how often a confident answer is given where none is warranted \\
Ident.\ AUROC & predicted against true resolvability & whether unresolvable cases are recognized \\
Surface error & median $|d_{\text{answer}}-d_{\text{ref}}|/d_{\text{ref}}$ inside the region, in disparity for stereo and depth for executed motion; multiplied by $100$ when reported in percent & how far the first physical surface lies from the stereo reference \\
Fooled & median relative error inside the region exceeds the error outside & whether region error exceeds surrounding error \\
\bottomrule
\end{tabular}
\end{table}

Table~\ref{tab:supp-metrics} defines the metrics used throughout this paper. The unresolvable
configurations that Section~\ref{sec:setup} retains occur in two forms. The first is a
view-tracked display observed within its aperture, which no baseline resolves. The second is a
mirror whose reflection of the operator stays outside the aperture or the frame under every permitted action. A planar mirror shares its content parallax with an opening of the same shape, so such a mirror leaves no trace. Each form also makes the physically present variant of
the same scene unresolvable, and on four mechanisms a pane joins them, because its image motion
equals that of an opening. Retaining these configurations keeps the abstention metrics
meaningful. A benchmark of resolvable cases alone would reward a method that answers
everywhere, because every scene would then be resolvable by some action.

\subsection{Comparison Methods and Ablations}
\label{sec:supp-baselines}

\begin{table}[htb]
\caption{\textbf{Three-mechanism synthetic benchmark.} Mean and sample standard deviation over $11$ seeds; chance is $0.333$. Committed: accuracy on answered scenes; Abst.\ res.: abstention on resolvable scenes; AUROC: identifiability AUROC. Section~\ref{sec:setup} defines CEA and FPCR. Ablations modify Intervene3D-sum; adequacy is disabled. A dash means unrecorded. }
\label{tab:phase1}
\centering
\footnotesize \setlength{\tabcolsep}{1.4pt}
\begin{tabular}{@{}lcccccc@{}}
\toprule
Method & CEA & Committed & Abstain & Abst.\ res. & AUROC & FPCR \\
\midrule
Single-view classifier    & $0.333{\pm}0.000$ & $0.333{\pm}0.000$ & $0\%$  & $0\%$ & $0.500{\pm}0.090$ & $0.000{\pm}0.000$ \\
Passive multi-view        & $0.622{\pm}0.030$ & $0.622{\pm}0.030$ & $0\%$  & $0\%$ & $0.725{\pm}0.072$ & $0.793{\pm}0.159$ \\
Next-best-view proxy & $0.027{\pm}0.017$ & $1.000{\pm}0.000$ & $97\%$ & --- & $0.890{\pm}0.044$ & $0.000{\pm}0.000$ \\
Random intervention       & $0.209{\pm}0.066$ & $1.000{\pm}0.000$ & $79\%$ & $67\%$ & $0.917{\pm}0.045$ & $0.000{\pm}0.000$ \\
Largest-translation & $0.396{\pm}0.050$ & $1.000{\pm}0.000$ & $60\%$ & $37\%$ & $0.970{\pm}0.015$ & $0.000{\pm}0.000$ \\
\groupband{7}{Ablations}
no hypothesis conditioning & $0.000{\pm}0.000$ & --- & $100\%$ & $100\%$ & $0.500{\pm}0.000$ & $0.000{\pm}0.000$ \\
noisy encoder & $0.430{\pm}0.056$ & $0.996{\pm}0.012$ & $57\%$ & --- & $1.000{\pm}0.000$ & $0.000{\pm}0.000$ \\
no abstention & $0.702{\pm}0.041$ & $0.702{\pm}0.041$ & $0\%$ & $0\%$ & $1.000{\pm}0.000$ & $0.764{\pm}0.159$ \\
\midrule
\oursrow \oursname{Intervene3D-sum} & $0.431{\pm}0.062$ & $\mathbf{1.000{\pm}0.000}$ & $57\%$ & $31\%$ & $1.000{\pm}0.000$ & $\mathbf{0.000{\pm}0.000}$ \\
\oursrow \oursname{Intervene3D-maximin} & $0.610{\pm}0.080$ & $\mathbf{1.000{\pm}0.000}$ & $39\%$ & $\mathbf{3\%}$ & $1.000{\pm}0.000$ & $\mathbf{0.000{\pm}0.000}$ \\
\bottomrule
\end{tabular}
\end{table}

The synthetic comparisons in Tables~\ref{tab:phase1} and~\ref{tab:phase2} use the following
methods. The classifiers use base-scene-disjoint training and evaluation splits.
The \emph{single-view classifier} is a softmax classifier on summary features of the initial landmarks, including position, depth, and visibility. \emph{Passive multi-view} is the same classifier with features that describe the response to one fixed, unchosen lateral translation. The \emph{next-best-view proxy} selects the largest predicted change from the initial geometry under the maximum-prior hypothesis. It is a generic proxy and implements no published next-best-view method. \emph{Random intervention} chooses uniformly among allowed non-null actions. \emph{Largest-translation} chooses the allowed action with the greatest translation magnitude and breaks ties by action order. Both then use the hypothesis-conditioned decision. The \emph{no-hypothesis-conditioning} ablation uses the direct transition for every hypothesis and retains its interface aperture for visibility. The \emph{noisy encoder} perturbs landmark coordinates by $0.6$~px and depths by $2\%$ relative Gaussian noise. All other components are those of Intervene3D-sum. The \emph{no-abstention} ablation disables all abstention checks and forces Intervene3D-sum to return a mechanism. Forced maximin applies the same change to the maximin selector.

On executed real motions, Table~\ref{tab:supp-motion-rules} also includes \emph{first frame},
which chooses the earliest accepted target frame. \emph{Largest-translation} chooses the
largest registered camera displacement. The other real-data rules are maximin, summed,
and random selection.

\subsection{Minimum Causal Resolving Baseline}
\label{sec:supp-mcrb}

\begin{proposition}[Minimum causal resolving baseline]\label{prop:mcrb}
Consider the pair $(H_D, H_E)$ under a pure lateral translation of baseline $B$, with a pinhole
camera of horizontal focal length $f$ pixels, and $B \ll Z$. Two scene points at depths
$Z_1 \neq Z_2$ then shift by approximately $fB/Z_i$, and they therefore produce a differential
parallax $fB\,|1/Z_1 - 1/Z_2|$. A static planar display, by contrast, produces none once the
screen-plane homography is compensated. Requiring that differential parallax to exceed the
separability threshold $\delta$ yields the smallest resolving baseline
\begin{equation}
\label{eq:mcrb}
B_{\min} \;>\; \frac{\delta}{f\left|1/Z_1 - 1/Z_2\right|},
\end{equation}
where $\delta$ coincides with $\epsilon$ whenever the motion term dominates the composite
distance of \eqref{eq:distance}.
\end{proposition}

The assumptions of Proposition~\ref{prop:mcrb} are restrictive and are checked before \eqref{eq:mcrb} is evaluated. A violated assumption therefore yields no prediction. The proposition covers neither view-tracked displays, which no baseline
resolves, nor mirrors, whose virtual scene carries genuine parallax.

\section{The Synthetic Benchmark}
This appendix reports the full selection results, the motion budget, the dielectric pane, mechanisms outside the family, and imperfect execution.

\subsection{Choosing the Intervention, in Full}
\label{sec:supp-phase2}

Table~\ref{tab:phase2} reports every objective on the three- and four-mechanism benchmarks. Every method
that abstains commits at an accuracy of $1.000$, on mirrors as elsewhere, so the objectives
differ in coverage. With four mechanisms, the pane predicts the image motion of an opening under every
action, and no action separates a direct scene from a transmission scene. Every method
therefore abstains on all direct and transmission scenes, and the commitments come from displays and mirrors. Maximin
abstains on $31.7\%$ of displays and $30.7\%$ of mirrors, against $26.2\%$ and $61.0\%$ for the summed objective. It abstains on no mirror that some action resolves. Forced to answer every scene,
maximin reaches a CEA of $0.610 \pm 0.015$ against $0.473 \pm 0.021$ for passive multi-view, a
paired gain of $+0.137$ ($t = 43.5$, $11$ of $11$ seeds).

\begin{table}[htb]
\caption{\textbf{Selection objectives with three and four mechanisms.} $11$ seeds, mean and sample standard deviation. Chance with four mechanisms is $0.250$. Mirror acc.: accuracy on answered mirror scenes; Abst.\ res.: abstention on resolvable scenes.}
\label{tab:phase2}
\centering
\footnotesize \setlength{\tabcolsep}{3pt}
\begin{tabular}{@{}lcccccc@{}}
\toprule
& \multicolumn{2}{c}{Mirror acc.} &
\multicolumn{4}{c}{Four mechanisms} \\
\cmidrule(lr){2-3}\cmidrule(l){4-7}
Objective & 3 mechanisms & 4 mechanisms & committed & abstains & abst.\ res. & CEA \\
\midrule
Single-view classifier & --- & --- & $0.250{\pm}0.000$ & $0\%$ & $0\%$ & $0.250{\pm}0.000$ \\
Passive multi-view     & --- & --- & $0.473{\pm}0.021$ & $0\%$ & $0\%$ & $0.473{\pm}0.021$ \\
Largest-translation           & --- & $1.000{\pm}0.000$ & $1.000{\pm}0.000$ & $75.5\%$ & $32.1\%$ & $0.245{\pm}0.026$ \\
Summed \eqref{eq:sum}  & $1.000{\pm}0.000$ & $1.000{\pm}0.000$ & $1.000{\pm}0.000$ & $71.8\%$ & $21.8\%$ & $0.282{\pm}0.019$ \\
\oursrow \oursname{Maximin} \eqref{eq:maximin} & $1.000{\pm}0.000$ & $1.000{\pm}0.000$ & $1.000{\pm}0.000$ & $65.6\%$ & $\mathbf{4.6\%}$ & $\mathbf{0.344{\pm}0.019}$ \\
\oursrow \oursname{Maximin}, forced choice & --- & $0.693{\pm}0.043$ & $0.610{\pm}0.015$ & $0\%$ & $0\%$ & $0.610{\pm}0.015$ \\
\bottomrule
\end{tabular}
\end{table}

\subsection{Motion Budget}
\label{sec:supp-budget}

\begin{figure}[htb]
\centering
\includegraphics[width=\linewidth]{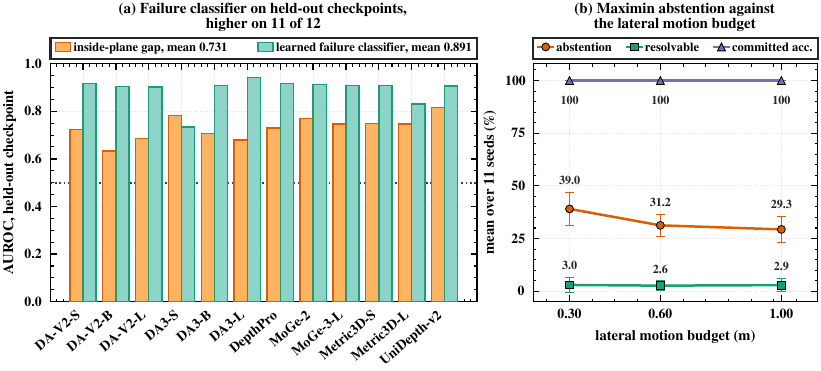}
\caption{\textbf{Failure classification and motion budget.} (a) Learned failure classifier and inside-plane gap on identical held-out-checkpoint folds (Appendix~\ref{sec:supp-learned}). (b) Maximin abstention on the three-mechanism benchmark at three lateral budgets, overall and on resolvable scenes; mean and sample standard deviation over $11$ seeds.}
\label{fig:transfer}
\end{figure}

Table~\ref{tab:supp-budget} widens the three-mechanism action set from $33$ actions within
$0.30$~m and $10^\circ$ to two larger budgets. These are $49$ actions within $0.60$~m and $15^\circ$, and $61$ actions within $1.00$~m laterally and $20^\circ$. Forward motion is capped at $0.60$~m, half the nearest interface distance. The base scenes and splits stay identical across budgets. A larger budget
resolves more scenes, and maximin abstention falls with it. Abstention on resolvable scenes stays near $3\%$, and committed accuracy stays at $1.000$. The summed and largest-translation selectors keep
abstaining on $29$ to $45\%$ of the resolvable scenes at every budget, and a random motion on
$67$ to $69\%$. Figure~\ref{fig:transfer}b plots the maximin
rows.

\begin{table}[htb]
\caption{\textbf{Three-mechanism benchmark under three motion budgets.} $11$ seeds, means. Every abstaining method commits at accuracy $1.000$. Forced maximin answers every scene.}
\label{tab:supp-budget}
\centering
\footnotesize \setlength{\tabcolsep}{3pt}
\begin{tabular}{@{}lccccccc@{}}
\toprule
& & \multicolumn{2}{c}{\oursname{Maximin}} & \multicolumn{2}{c}{Summed} & \hdr{Largest-\\translation} & \oursname{Forced maximin} \\
\cmidrule(lr){3-4}\cmidrule(lr){5-6}\cmidrule(lr){7-7}\cmidrule(l){8-8}
Budget & Resolvable & abstains & abst.\ res. & abstains & abst.\ res. & abst.\ res. & CEA \\
\midrule
$0.30$~m, $33$ actions & $62.8\%$ & $39.0\%$ & $3.0\%$ & $56.9\%$ & $30.9\%$ & $36.7\%$ & $0.799$ \\
$0.60$~m, $49$ actions & $70.6\%$ & $31.2\%$ & $2.6\%$ & $53.8\%$ & $34.6\%$ & $39.9\%$ & $0.844$ \\
$1.00$~m, $61$ actions & $72.8\%$ & $29.3\%$ & $2.9\%$ & $48.1\%$ & $28.7\%$ & $45.3\%$ & $0.855$ \\
\bottomrule
\end{tabular}
\end{table}

\subsection{Glass as a Dielectric Interface}
\label{sec:supp-glass}

The transmitting pane in the four-mechanism benchmark forms the same image as an opening, so motion can never
separate the two. A real pane also reflects. The dielectric variant keeps the transmitted
content at its apparent point and adds the pane's weak reflection of the operator markers,
visible when the Fresnel reflectance at the viewing angle times the marker's contrast reaches
$2\%$. Brightness enters only through that visibility flag. Every choice is fixed before any run. Table~\ref{tab:supp-glass} compares the two pane models with four mechanisms. The dielectric pane
makes $48.1\%$ of direct scenes and $22.7\%$ of transmission scenes resolvable, since the pane
now shows a reflection that an opening lacks. Mirrors lose resolvability for the same reason, because a pane
and a mirror reflect the same markers and differ only in brightness, which the model does not
use as evidence. Committed accuracy stays at $1.000$ on all four mechanisms, and maximin keeps
its abstention on resolvable scenes near $5\%$ against $25.2\%$ for the summed objective. A
sensitivity check run after these results, on the generated scenes of three seeds, fixes every
marker's contrast instead of drawing it. At a contrast of $0.25$ or more no transmission scene is
resolvable, and at $0.50$ or more no mirror either, because the pane then shows every marker that the mirror shows. The pane is told apart from a mirror only through dim markers that the
pane hides, so the transmission result depends on the spread of marker contrast.

\begin{table}[htb]
\caption{\textbf{Four-mechanism benchmark with transmitting and dielectric panes.} Maximin, $11$ seeds. Resolvable shares and abstention per mechanism.}
\label{tab:supp-glass}
\centering
\footnotesize \setlength{\tabcolsep}{4pt}
\begin{tabular}{@{}lcccc@{}}
\toprule
& \multicolumn{2}{c}{Transmitting pane} & \multicolumn{2}{c}{Dielectric pane} \\
\cmidrule(lr){2-3}\cmidrule(l){4-5}
Mechanism & resolvable & abstains & resolvable & abstains \\
\midrule
Direct       & $0.0\%$  & $100\%$  & $48.1\%$ & $55.9\%$ \\
Emissive     & $74.8\%$ & $31.7\%$ & $74.4\%$ & $30.1\%$ \\
Reflection   & $69.3\%$ & $30.7\%$ & $25.2\%$ & $74.8\%$ \\
Transmission & $0.0\%$  & $100\%$  & $22.7\%$ & $77.3\%$ \\
\midrule
All, CEA / abst.\ res. & \multicolumn{2}{c}{$0.344$ / $4.6\%$} & \multicolumn{2}{c}{$0.405$ / $4.9\%$} \\
Summed, CEA / abst.\ res. & \multicolumn{2}{c}{$0.282$ / $21.8\%$} & \multicolumn{2}{c}{$0.319$ / $25.2\%$} \\
\bottomrule
\end{tabular}
\end{table}

\subsection{Mechanisms Outside the Family}
\label{sec:supp-e5}

Every test base scene of a four-mechanism benchmark carries four held-out variants besides its in-family variants. The held-out variants share the scene's interface, reference camera, content and operator markers. They are a spherical mirror whose sag stops before the operator's reflection enters the frame
at $C_0$, the same mirror over the full sag range, a cylindrical display, and a water-like
surface that reflects half of the landmarks and refracts the others. Reference views match to
below $2\times10^{-10}$~px wherever the operator stays out of frame. The adequacy threshold of
$2$~px was fixed before any result. Table~\ref{tab:family} reports wrong commitments with the test off and on. On in-family scenes, the test costs nothing with an exact encoder and
$0.025$ of CEA with an encoder perturbed by $0.6$~px.

\begin{table}[htb]
\caption{\textbf{Mechanisms outside the four-member family.} Maximin selection, $11$ seeds. Wrong commitments count held-out scenes on which the system commits to a family member, every one of which is wrong. The adequacy test abstains when the smallest residual exceeds $2$~px, a value fixed before any run, and \emph{Test fires} gives its rate with an exact encoder. Sag is relative to the half-diagonal of a mirror or the half-width of a panel, as a median. $^\dagger$Reference views coincide on $48\%$ of these scenes, because a strongly curved mirror reflects the operator into view at $C_0$.}
\label{tab:family}
\centering
\footnotesize \setlength{\tabcolsep}{3.2pt}
\begin{tabular}{@{}lccccc@{}}
\toprule
& & \multicolumn{2}{c}{Exact encoder} & \multicolumn{2}{c}{Noisy encoder} \\
\cmidrule(lr){3-4}\cmidrule(l){5-6}
Held-out mechanism & Test fires & test off & test on & test off & test on \\
\midrule
Spherical mirror, sag $0.012$ & $39.8\%$ & $0.618$ & $0.397$ & $0.641$ & $0.331$ \\
Spherical mirror, full sag range$^\dagger$, sag $0.051$ & $71.5\%$ & $0.568$ & $\mathbf{0.124}$ & $0.582$ & $\mathbf{0.084}$ \\
Cylindrical display, sag $0.057$ & $43.2\%$ & $0.602$ & $0.432$ & $0.595$ & $0.378$ \\
Water-like surface, reflection and refraction & $4.2\%$ & $0.656$ & $0.643$ & $0.656$ & $0.600$ \\
Three primary mechanisms, pooled & $29.1\%$ & $0.625$ & $0.491$ & $0.631$ & $0.436$ \\
\midrule
Spherical mirror, family with a known-sag member & --- & $0.000$ & $0.000$ & --- & --- \\
\bottomrule
\end{tabular}
\end{table}

With an exact encoder, the true in-family hypothesis leaves a zero residual, so the smallest
residual ranks every held-out scene above every in-family scene and the detection AUROC is
$1.000$ by construction. With an encoder perturbed by $0.6$~px, the AUROC is $0.801$ for the
spherical mirror, $0.696$ for the cylindrical display, $0.548$ for the water-like surface and
$0.904$ over the full sag range. Detection grows with curvature. For the spherical mirror, the
test fires on $13\%$ of scenes below a sag of $0.01$, $46\%$ between $0.01$ and $0.03$, and
$77\%$ to $100\%$ above, where it removes every wrong commitment. The cylindrical display needs
a sag above $0.1$ before the test fires on most scenes. The water-like surface leaves a median smallest residual of
$0.18$~px, below any useful threshold, and its commitments go to $H_R$ in $65\%$ of scenes.

A fifth member, a spherical mirror with the scene's sag and its closed-form transition,
answers correctly on every spherical mirror it commits to. It reaches a CEA of $0.896$ with $10.4\%$ abstention, and $0.928$ with $7.2\%$ over the full sag range. Carrying a nominal sag of $0.05$
instead, it commits wrongly on $43.0\%$ of those scenes. On in-family scenes the fifth member
lowers the CEA of maximin by $0.019$.

\subsection{Imperfect Execution}
\label{sec:supp-noise}

A commanded motion rarely arrives exactly, and a selector that depends on the exact motion would lose its advantage in deployment. Table~\ref{tab:noise} therefore reports the effect of executing a chosen action imperfectly, at $\pm 1$~cm and $\pm 0.5^{\circ}$, with the passive baseline perturbed in the same way. Intervene3D-sum loses $5.0\%$ of its CEA, the
largest-translation rule $4.4\%$ and passive multi-view $3.8\%$, whereas the next-best-view proxy
loses $21.2\%$ and a random choice $25.6\%$. Intervene3D therefore degrades about as much as the
largest-translation rule and far less than the next-best-view proxy and the random choice. Maximin selection
was not run under perturbed execution.

\begin{table}[htb]
\caption{\textbf{CEA under clean and perturbed execution.} Errors of $\pm 1$~cm and $\pm 0.5^{\circ}$, with the paired $t$ statistic over the same eleven seeds.}
\label{tab:noise}
\centering
\footnotesize \setlength{\tabcolsep}{4pt}
\begin{tabular}{@{}lcccc@{}}
\toprule
Method & Clean CEA & Perturbed CEA & Change & $t$ \\
\midrule
Passive multi-view & $0.622$ & $0.598$ & $-3.8\%$ & $-1.92$ \\
Next-best-view proxy   & $0.027$ & $0.021$ & $-21.2\%$ & $-2.04$ \\
Random intervention & $0.209$ & $0.156$ & $-25.6\%$ & $-5.04$ \\
Largest-translation        & $0.396$ & $0.378$ & $-4.4\%$  & $-3.88$ \\
\oursrow \oursname{Intervene3D-sum} & $0.431$ & $0.410$ & $-5.0\%$ & $-5.20$ \\
\bottomrule
\end{tabular}
\end{table}

\section{Real Data}
This appendix details the executed-motion and stereo experiments and the per-checkpoint and qualitative results. It then covers real mirrors, ordinal labels, mirrors in video, and learned failure prediction. It ends with scene-level statistics, further detectors, and selection under restricted action sets.

\subsection{Executed Motions on the Illusion Benchmark}
\label{sec:supp-e1}

\begin{figure}[htb]
\centering
\includegraphics[width=\linewidth]{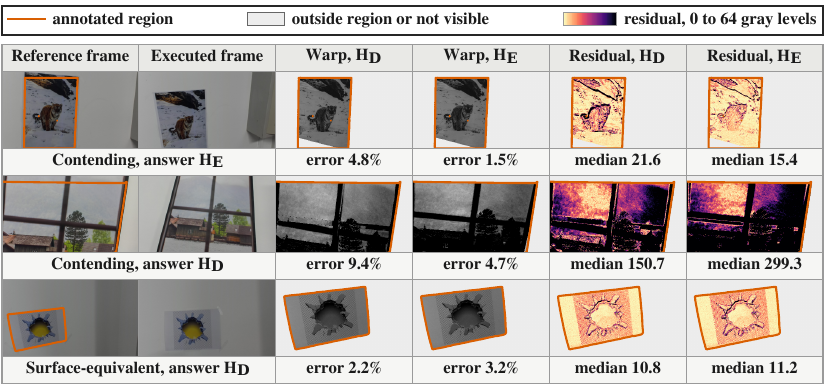}
\caption{\textbf{Three executed-motion decisions for DA-V2-Small.} The rows are the median contending regions answered $H_E$ and $H_D$ and the median surface-equivalent region. The columns are the reference and executed frames, followed by the warp and the absolute photometric residual under each hypothesis on a fixed 0 to 64 gray-level scale. Values below the warps and residual maps are surface errors and median residuals in gray levels.}
\label{fig:real-motion}
\end{figure}

The illusion benchmark \citep{illusion2025} photographs most static scenes from several
camera positions with a calibrated stereo rig, and its reference depth makes those positions
recoverable. Every admissible frame of the $53$ scenes with at least three admissible frames serves as a reference, which gives $196$ regions covering at least $3\%$ of the image. Outside the regions the scenes
offer too few features, so poses use features from the whole image, matched against the
reference depth, which lies on physical surfaces. Of the $906$ ordered pairs, $528$ pass the
acceptance thresholds, with a median of $878$ inliers, a median reprojection error of $0.80$~px
and a forward-backward cycle error of $0.85^\circ$ and $1.3$~cm. Of the $196$ references, $135$
keep at least one accepted target and $61$ keep one within $10^\circ$ and $0.30$~m. The real
frames differ by $20$ to $25^\circ$ of rotation, so each accepted pair supplies one action, and the envelope subset is reported separately. Table~\ref{tab:motion} reports the decision,
Figure~\ref{fig:real-motion} follows three regions through it, and
Table~\ref{tab:supp-motion-rules} compares five rules for choosing the executed motion.

\begin{table}[htb]
\caption{\textbf{Mechanism accuracy under five rules for choosing the motion.} All regions of the executed real motions, twelve checkpoints, $135$ regions each. Accuracy over all regions counts an answer of $H_D$ as wrong also where both hypotheses agree on the surface. With two hypotheses, maximin and the summed objective select the same motion. The envelope is that of Table~\ref{tab:motion}.}
\label{tab:supp-motion-rules}
\centering
\footnotesize \setlength{\tabcolsep}{4pt}
\begin{tabular}{@{}lccccc@{}}
\toprule
& \oursname{Maximin} & Summed & Random & \hdr{Largest-\\translation} & First frame \\
\midrule
Every accepted motion & $69.1\%$ & $69.1\%$ & $68.9\%$ & $68.2\%$ & $69.6\%$ \\
Within the envelope & $65.7\%$ & $65.7\%$ & $66.9\%$ & $68.0\%$ & $69.2\%$ \\
\bottomrule
\end{tabular}
\end{table}

In the first row of Figure~\ref{fig:real-motion}, the decision lowers surface error from $4.8\%$ to $1.5\%$.
In the second row the photometric fit is degenerate: gain is $3.97$ and both residuals exceed
$150$ gray levels. The decision returns $H_D$ although the plane would have lower surface
error. The third row is surface-equivalent and its answer is $H_D$.

Several outcomes follow from properties of the data. In $957$ of the $1{,}620$
region--checkpoint combinations, the hypotheses place the surface within $2\%$; their surface
errors differ by a median of $0.55$ percentage points. Among the $663$ contending combinations,
every accepted motion separates the pair by at least $1$~px in $96.4\%$, which explains the small effect of selection. Appendix~\ref{sec:supp-sweep} tests selection where three hypotheses contend and the action set is restricted. Within the envelope, $6$ of $732$ combinations abstain, with
separabilities from $0.52$ to $0.93$~px. The commitment threshold $\tau$ cannot bind with two
hypotheses under a uniform prior, since the larger posterior is at least $0.5$, and the real
protocol commits to the smallest residual. Both the answer and the predictor use reference
depth outside the region for their alignment, so the comparison between them is like for like.

\subsection{One Stereo Baseline, by Category}
\label{sec:supp-stereo}

\begin{figure}[htb]
\centering
\includegraphics[width=\linewidth]{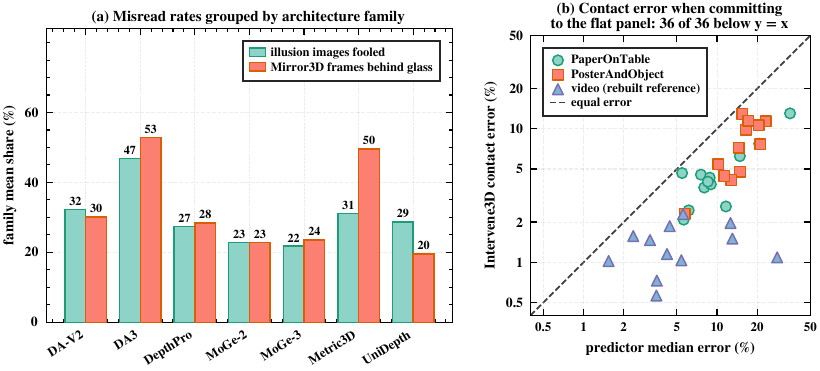}
\caption{\textbf{Failure rates and surface error at commitment.} (a) Fooled share on illusion images and behind-glass share on Mirror3D, averaged within architecture family. (b) Surface error at $H_E$ commitments for printed sheets, posters, and monitors in the frame; one point per checkpoint and category, logarithmic axes. The $36$ points cover three categories; monitors filling the frame are omitted. Monitor-in-the-frame results use rebuilt reference disparity. The dashed line denotes equal error.}
\label{fig:supp-extended}
\end{figure}

Figure~\ref{fig:supp-extended}b plots, per checkpoint and category, the surface error of a
commitment to $H_E$ against the predictor's error on the same images. All $36$ points lie below
equal error. Table~\ref{tab:supp-identifiable} compares separability between fooled and correctly predicted images within each category, with the threshold and decision of the main protocol, in an analysis made after the
results. In the three categories whose ring plane lies on the panel, fooled images are $2.2$ to
$8.9$ times more often separable than correctly predicted ones, and where a fooled image is
separable the system commits and lowers the surface error by $40$ to $76\%$. Over all four categories, two thirds of fooled images are not separable from a single stereo baseline, and on those the system abstains, as identifiability requires. Table~\ref{tab:supp-category} completes Table~\ref{tab:category} with the error ratio
and the median separability.

\begin{table}[htb]
\caption{\textbf{Separability and failure within category.} Twelve checkpoints, $\epsilon = 2.0$ gray levels, analyzed after the results. The last two columns compare the surface error of the answer and of the predictor on fooled images where the system commits. $^\ddagger$The ring plane lies behind the screen in this category.}
\label{tab:supp-identifiable}
\centering
\footnotesize \setlength{\tabcolsep}{3pt}
\begin{tabular}{@{}lccccc@{}}
\toprule
Category & \hdr{Fooled\\pairs} & \hdr{$\Delta \geq \epsilon$,\\fooled} & \hdr{$\Delta \geq \epsilon$,\\correct} & \hdr{Surface error,\\\textcolor{oursink}{answer} / \textcolor{baseink}{predictor}} & \hdr{Answer\\lower} \\
\midrule
Printed sheet on a table & $498$ & $36.5\%$ & $16.3\%$ & $0.053$ / $0.120$ & $72.0\%$ \\
Poster beside an object & $844$ & $24.8\%$ & $8.2\%$ & $0.090$ / $0.149$ & $68.4\%$ \\
Monitor in the frame & $138$ & $56.5\%$ & $6.4\%$ & $0.011$ / $0.047$ & $98.7\%$ \\
\midrule
Monitor filling the frame$^\ddagger$ & $137$ & $41.6\%$ & $69.9\%$ & $0.440$ / $0.440$ & $3.5\%$ \\
\bottomrule
\end{tabular}
\end{table}

\begin{table}[htb]
\caption{\textbf{Separability by category.} The categories of Table~\ref{tab:category}, completed with the ratio of the error inside the region to the error outside and the median separability, with the plane fitted inside the region in brackets.}
\label{tab:supp-category}
\centering
\footnotesize \setlength{\tabcolsep}{5pt}
\begin{tabular}{@{}lccc@{}}
\toprule
Category & Error in / out & Median $\Delta_{DE}$ & $\Delta_{DE} < \epsilon$ \\
\midrule
Printed sheet on a table & $0.96\times$ & $1.24$ [$0.84$] & $74.1\%$ [$84.6\%$] \\
Poster beside an object & $1.25\times$ & $1.00$ [$0.71$] & $82.0\%$ [$87.7\%$] \\
Monitor in the frame & $0.39\times$ & $0.85$ [$0.31$] & $87.8\%$ [$97.5\%$] \\
Monitor filling the frame & $0.26\times$ & $2.77$ [$0.51$] & $33.3\%$ [$95.2\%$] \\
\midrule
All admissible & $0.66\times$ & $1.35$ [$0.54$] & $69.8\%$ [$91.3\%$] \\
\bottomrule
\end{tabular}
\end{table}

\subsection{Per-Checkpoint Results}
\label{sec:supp-models}

Table~\ref{tab:models} lists every checkpoint with its parameter count and family. Parameter
count carries no signal. The smallest checkpoints hold $24.8$M parameters and the largest
$952.0$M, a range of $38\times$. DepthPro at the top of that range is fooled on $27.5\%$ of
illusion images against $28.2\%$ for DA-V2-Small at the bottom. DA3-Small fails most often on
illusions and Metric3D-Small most often on mirrors, and the two rankings differ, so no single
checkpoint dominates both failure modes. The executed motion lowers the median surface error
on every checkpoint. Figure~\ref{fig:supp-extended}a averages within family.

\begin{table}[htb]
\caption{\textbf{Per-checkpoint results.} Fooled shares use the $404$ admissible illusion images, behind-glass shares the $123$ Mirror3D frames, and the surface errors the $135$ regions under executed motion. The last two columns give the leave-one-model-out AUROC of the learned failure classifier and of the strongest single feature, the gap between the prediction and a plane fitted inside the region. $^\ast$MoGe-2 and MoGe-3 report no depth on part of some regions; such images count as not fooled.}
\label{tab:models}
\centering
\footnotesize \setlength{\tabcolsep}{2.4pt}
\begin{tabular}{@{}llrcccccc@{}}
\toprule
& & & Illusion & Mirror3D & \multicolumn{2}{c}{Executed motion} & \multicolumn{2}{c}{AUROC} \\
\cmidrule(lr){4-4}\cmidrule(lr){5-5}\cmidrule(lr){6-7}\cmidrule(l){8-9}
Checkpoint & Family & Params & Fooled & Behind glass & \textcolor{baseink}{Predictor} & \textcolor{oursink}{Answer} & Learned & Single \\
\midrule
DA-V2-Small     & Depth-Anything-V2 & $24.8$M  & $28.2\%$ & $32.5\%$ & $5.23\%$ & $3.43\%$ & $0.917$ & $0.722$ \\
DA-V2-Base      & Depth-Anything-V2 & $97.5$M  & $32.7\%$ & $30.1\%$ & $3.75\%$ & $2.78\%$ & $0.905$ & $0.634$ \\
DA-V2-Large     & Depth-Anything-V2 & $335.3$M & $35.9\%$ & $27.6\%$ & $3.45\%$ & $2.12\%$ & $0.902$ & $0.685$ \\
DA3-Small       & Depth-Anything-3  & $24.8$M  & $\mathbf{68.3\%}$ & $59.3\%$ & $8.20\%$ & $5.30\%$ & $0.735$ & $0.783$ \\
DA3-Base        & Depth-Anything-3  & $97.5$M  & $37.1\%$ & $56.9\%$ & $5.22\%$ & $3.64\%$ & $0.908$ & $0.706$ \\
DA3-Large       & Depth-Anything-3  & $410.9$M & $34.9\%$ & $42.3\%$ & $4.83\%$ & $2.43\%$ & $0.942$ & $0.680$ \\
DepthPro        & DepthPro          & $952.0$M & $27.5\%$ & $28.5\%$ & $3.48\%$ & $2.52\%$ & $0.916$ & $0.729$ \\
MoGe-2$^\ast$   & MoGe-2            & $330.9$M & $22.8\%$ & $22.8\%$ & $3.28\%$ & $1.87\%$ & $0.912$ & $0.770$ \\
MoGe-3-ViT-L$^\ast$ & MoGe-3        & $330.9$M & $21.8\%$ & $23.6\%$ & $2.90\%$ & $2.00\%$ & $0.908$ & $0.747$ \\
Metric3D-Small  & Metric3D          & $37.5$M  & $32.9\%$ & $\mathbf{66.7\%}$ & $6.26\%$ & $4.35\%$ & $0.909$ & $0.749$ \\
Metric3D-Large  & Metric3D          & $343.0$M & $29.5\%$ & $32.5\%$ & $4.14\%$ & $3.64\%$ & $0.831$ & $0.747$ \\
UniDepth-v2     & UniDepth          & $353.8$M & $28.7\%$ & $19.5\%$ & $4.13\%$ & $3.03\%$ & $0.906$ & $0.815$ \\
\midrule
\multicolumn{3}{@{}l}{\textbf{Mean across checkpoints}} & $\mathbf{33.4\%}$ & $\mathbf{36.9\%}$ & --- & --- & $\mathbf{0.891}$ & $0.731$ \\
\multicolumn{3}{@{}l}{\textbf{Pooled median surface error}} & --- & --- & $4.35\%$ & $\mathbf{2.98\%}$ & --- & --- \\
\bottomrule
\end{tabular}
\end{table}

\subsection{Qualitative Results on Stereo Photographs}
\label{sec:supp-qual}

\begin{figure}[htb]
\centering
\includegraphics[width=\linewidth]{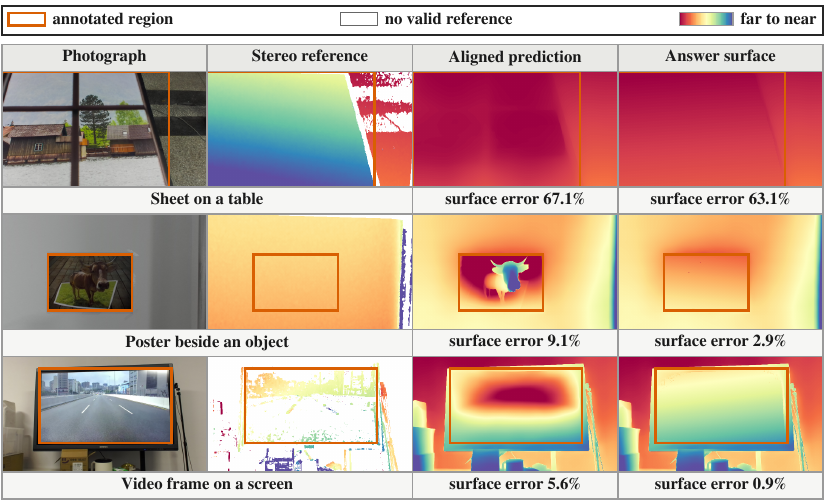}
\caption{\textbf{Decisions on three photographs.} Each row is the median case by surface-error reduction among admissible, fooled images answered $H_E$ by DA-V2-Small. The columns are the photograph with the annotated region outlined, the stereo reference, the aligned prediction and the answered surface, on one color scale per row. The surface errors of the prediction and of the answer are printed below the two right columns. The video row uses the rebuilt stereo reference.}
\label{fig:qualitative}
\end{figure}

\begin{figure}[htb]
\centering
\includegraphics[width=\linewidth]{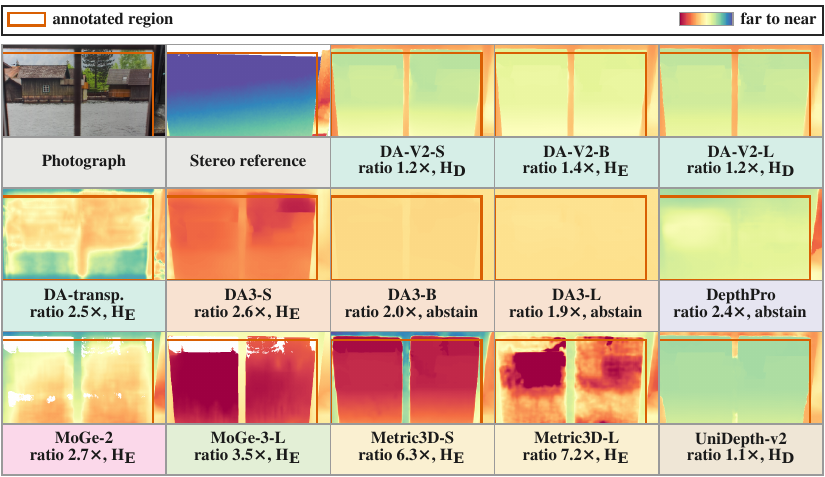}
\caption{\textbf{Thirteen checkpoints on one poster.} A poster beside a window frame is shown with its stereo reference and thirteen aligned predictions on one color scale. Each panel is labelled with the inside-to-outside surface-error ratio and the decision at $\epsilon=2.0$ gray levels. MoGe-2 and MoGe-3-L leave part of the region without depth (white), and their ratio uses the pixels with estimated depth.}
\label{fig:model-grid}
\end{figure}

Figure~\ref{fig:qualitative} shows one photograph per category. Each is the median case by
surface-error reduction among admissible, fooled images on which DA-V2-Small commits to $H_E$,
selected by a rule written before any figure was rendered. The median sheet shows how little a
commitment can gain when the ring lies on a different surface from the sheet. The region covers
$73\%$ of the frame and reaches its borders, and the plane improves the surface error only from
$67.1\%$ to $63.1\%$. The poster and the video frame improve from $9.1\%$ to $2.9\%$ and from
$5.6\%$ to $0.9\%$. Figure~\ref{fig:model-grid} places all thirteen checkpoints, the
transparency variant included, on one photograph of a poster beside a window frame. The DA-transp. panel shows a Depth Anything variant fine-tuned for transparent surfaces. It returns metric depth only with a depth prompt, so its alignment fails under the single-image protocol and it stays outside the twelve-checkpoint evaluation (Section~\ref{sec:setup}). It is included as the one checkpoint trained for such surfaces, and on this poster its error inside the region is still $2.5$ times its error outside. Across the thirteen checkpoints, the decision commits to the panel for seven, to the prediction for three, and abstains for three.

\subsection{Real Mirrors, with Annotated First Physical Surfaces}
\label{sec:supp-mirrors}

The illusion benchmark contains no photograph of a physical mirror. Mirror3D
\citep{mirror3d2021} annotates real mirrors with an instance mask and a three-dimensional
mirror plane, layered on NYUv2 \citep{nyuv22012}. The annotated plane serves as ground truth here because a depth sensor does not. A structured-light sensor returns
either no signal or unreliable depth on a mirror, and the human-placed plane therefore remains
the only statement of where the first physical surface lies. Scale and shift are fitted outside the annotated region against reference depth and applied inside it, in inverse-depth space. The predictors return relative inverse depth, and conversion to meters would require the very scale that the annotation leaves open on a mirror. The released annotations carry a $5\%$ center crop, and the principal point is shifted to match.

\begin{figure}[htb]
\centering
\includegraphics[width=\linewidth]{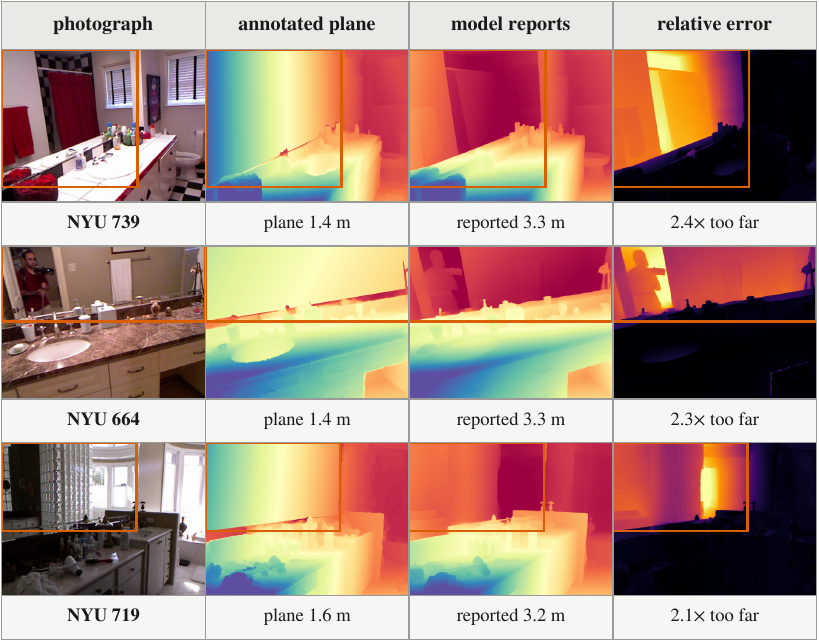}
\caption{\textbf{Individual Mirror3D frames for DA-V2-Small.} Columns show the image, annotated mirror plane, aligned network estimate, and relative error. The plane is outlined in every panel. Row labels give plane depth in meters and the ratio of estimated depth to plane depth.}
\label{fig:supp-mirrors}
\end{figure}

Table~\ref{tab:mirrors} reports the outcome over $123$ annotated frames. The primary statistic
is the share of frames whose predicted surface lies behind the mirror plane, a direct statement about the first surface. It averages $36.9\%$ and reaches two thirds for Metric3D-Small. The
fooled label of the stereo protocol, a larger error inside the mirror than outside, holds on
$69.1\%$ of frames on average, and the error inside reaches $5.47$ times the error outside.
Placing the surface behind the wall reproduces the reflected room, which is the same failure
that a display produces with a depicted scene. NYUv2 provides no camera poses, so these frames
measure the predictors and not the decision. Figure~\ref{fig:supp-mirrors} presents individual
frames.

\begin{table}[htb]
\caption{\textbf{Twelve checkpoints on real mirrors.} $123$ annotated frames \citep{mirror3d2021}. The columns give the fooled share, the ratio of the relative error inside to outside the mirror region, and the share of frames placing the first surface behind the glass.}
\label{tab:mirrors}
\centering
\footnotesize \setlength{\tabcolsep}{4pt}
\begin{tabular}{@{}llccc@{}}
\toprule
Checkpoint & Family & Fooled & Error inside / outside & Surface behind glass \\
\midrule
DA-V2-Small     & Depth-Anything-V2 & $65.0\%$ & $1.72\times$ & $32.5\%$ \\
DA-V2-Base      & Depth-Anything-V2 & $71.5\%$ & $2.00\times$ & $30.1\%$ \\
DA-V2-Large     & Depth-Anything-V2 & $65.0\%$ & $1.82\times$ & $27.6\%$ \\
DA3-Small       & Depth-Anything-3  & $\mathbf{79.7\%}$ & $2.85\times$ & $59.3\%$ \\
DA3-Base        & Depth-Anything-3  & $76.4\%$ & $3.38\times$ & $56.9\%$ \\
DA3-Large       & Depth-Anything-3  & $75.6\%$ & $2.57\times$ & $42.3\%$ \\
DepthPro        & DepthPro          & $65.0\%$ & $1.91\times$ & $28.5\%$ \\
MoGe-2          & MoGe-2            & $74.8\%$ & $2.19\times$ & $22.8\%$ \\
MoGe-3-ViT-L    & MoGe-3            & $73.2\%$ & $2.09\times$ & $23.6\%$ \\
Metric3D-Small  & Metric3D          & $78.9\%$ & $\mathbf{5.47\times}$ & $\mathbf{66.7\%}$ \\
Metric3D-Large  & Metric3D          & $41.5\%$ & $0.82\times$ & $32.5\%$ \\
UniDepth-v2     & UniDepth          & $62.6\%$ & $1.93\times$ & $19.5\%$ \\
\midrule
\textbf{Mean} & & $\mathbf{69.1\%}$ & --- & $\mathbf{36.9\%}$ \\
\bottomrule
\end{tabular}
\end{table}

\subsection{Second Dataset with Ordinal Supervision}
\label{sec:supp-layered}

The second dataset, LayeredDepth, provides ordinal relations between point pairs rather than dense depth,
together with a per-point layer index authored independently of the present work. The ordinal
form suits the question here, because a relation between two points holds under the scale and
shift ambiguity that a monocular predictor leaves open. Splitting the relations by layer index
then separates pairs that lie on a first visible surface from pairs that lie behind glass, with
no quantity of ours entering the split. DA-V2-Small and DepthPro answer the single-layer pairs well and are less accurate on the multi-layer pairs, at $0.848$ against $0.737$ for DA-V2-Small and
$0.843$ against $0.669$ for DepthPro (Table~\ref{tab:supp-layered}). The two networks agree closely on the surfaces they can see, at first-surface accuracies of $0.847$ and $0.842$, but diverge on the surfaces behind an interface, where the ambiguity under study arises.

\begin{table}[htb]
\caption{\textbf{LayeredDepth validation} \citep{layereddepth2025}\textbf{.} Ordinal accuracy split by the per-point layer index into single-layer and multi-layer point pairs.}
\label{tab:supp-layered}
\centering
\footnotesize \setlength{\tabcolsep}{4pt}
\begin{tabular}{@{}lccccc@{}}
\toprule
Checkpoint & Single-layer & Multi-layer & Gap & First surface & Through glass \\
\midrule
DA-V2-Small & $0.848$ & $0.737$ & $\mathbf{+0.111}$ & $0.847$ & $0.784$ \\
DepthPro    & $0.843$ & $0.669$ & $\mathbf{+0.175}$ & $0.842$ & $0.741$ \\
\bottomrule
\end{tabular}
\end{table}

\subsection{Mirrors in Real Video}
\label{sec:supp-e8}

ScanNet++ \citep{scannetpp2023} provides handheld video with registered poses and a laser-scanned mesh. The mirror plane is taken from this mesh for $30$ scenes and $1{,}433$ mirror regions. The regions are evaluated for all twelve checkpoints with the hypothesis set $\{H_D, H_E, H_R\}$, which gives $17{,}196$ region--checkpoint combinations. The operator's reflection is not annotated, so no operator cue is
available, and without one a planar mirror and the opening it imitates predict the same images
under every motion. The protocol, written before any run, therefore predicted abstention on the
pair of $H_D$ and $H_R$. The system abstains on $75.8\%$ of combinations, and the identifiability of that pair is zero in every abstention. It commits only where all hypotheses place the surface within $2\%$, so its answers
match the prediction there, at a median surface error of $0.036$ for both. A surface is counted as confidently wrong when it is committed and lies more than $10\%$ from the mesh plane. In a post hoc analysis, confidently wrong surfaces fall from $35.3\%$ to $5.0\%$ of
combinations, entirely through abstention. Table~\ref{tab:supp-e8} shows that the share is lower for all twelve checkpoints, and it is also lower in all $30$ scenes.
The experiment confirms the identifiability analysis on real mirrors. Without an operator cue, no motion can separate these mirrors from the openings they imitate, and the system declines these regions instead of committing to a wrong surface.

\begin{table}[htb]
\caption{\textbf{ScanNet++ mirrors per checkpoint.} $30$ scenes and $1{,}433$ mirror regions per checkpoint, hypothesis set $\{H_D,H_E,H_R\}$, no operator cue. A surface is confidently wrong when it is committed and lies more than $10\%$ from the mesh plane; the network always commits.}
\label{tab:supp-e8}
\centering
\footnotesize \setlength{\tabcolsep}{5pt}
\begin{tabular}{@{}lccc@{}}
\toprule
& & \multicolumn{2}{c}{Confidently wrong surfaces} \\
\cmidrule(l){3-4}
Checkpoint & Abstention & \textcolor{baseink}{Network} & \textcolor{oursink}{Intervene3D} \\
\midrule
DA-V2-Small     & $76.2\%$ & $32.0\%$ & $4.4\%$ \\
DA-V2-Base      & $73.6\%$ & $24.9\%$ & $3.8\%$ \\
DA-V2-Large     & $74.3\%$ & $23.6\%$ & $3.3\%$ \\
DA3-Small       & $87.2\%$ & $51.6\%$ & $2.9\%$ \\
DA3-Base        & $81.5\%$ & $40.0\%$ & $2.3\%$ \\
DA3-Large       & $78.0\%$ & $33.4\%$ & $2.0\%$ \\
DepthPro        & $77.0\%$ & $31.3\%$ & $4.2\%$ \\
MoGe-2          & $76.6\%$ & $21.8\%$ & $1.3\%$ \\
MoGe-3-ViT-L    & $72.4\%$ & $19.4\%$ & $1.3\%$ \\
Metric3D-Small  & $79.4\%$ & $49.2\%$ & $5.4\%$ \\
Metric3D-Large  & $59.5\%$ & $74.8\%$ & $26.9\%$ \\
UniDepth-v2     & $73.8\%$ & $21.8\%$ & $1.9\%$ \\
\midrule
\textbf{Pooled} & $75.8\%$ & $35.3\%$ & $\mathbf{5.0\%}$ \\
\bottomrule
\end{tabular}
\end{table}

\subsection{Learned Failure Prediction and Consistency Checks}
\label{sec:supp-learned}

A gradient-boosted failure classifier uses per-image quantities to predict the fooled label.
Its inputs are the photometric separability, the residuals of both hypotheses, texture, region
size, the disparity gap, the alignment quality outside the region, the gap between the
prediction and a plane fitted inside the region, the predictor's confidence, and test-time
augmentation disagreement. The boosted models stay at depth $3$ with at least $40$ samples per
leaf, and nothing is tuned on the evaluation set. Table~\ref{tab:failure-splits} separates
evaluation splits and category pools. With only the checkpoint held out, the same images
remain in training under other checkpoints. Holding out their scenes as well tests an unseen
network on unseen scenes. The classifier outperforms the inside-plane-gap feature on $11$ of $12$
checkpoint folds (Figure~\ref{fig:transfer}a, Table~\ref{tab:models}). Lower within-category
AUROCs show that part of the pooled score reflects category mix.

\begin{table}[htb]
\caption{\textbf{Failure-classification AUROC by evaluation split.} Each row specifies its category pool. A dash means no corresponding single-feature value is reported.}
\label{tab:failure-splits}
\centering
\footnotesize\setlength{\tabcolsep}{4pt}
\begin{tabular}{@{}llcc@{}}
\toprule
Held out & Evaluation pool & Classifier & Inside-plane gap \\
\midrule
Checkpoint & All categories & $0.891$ & $0.731$ \\
Checkpoint and images & All categories & $0.870$ & --- \\
Checkpoint and scenes & All categories & $0.859$ & --- \\
Scenes, within category & Printed sheet on a table & $0.70$ & $0.64$ \\
Scenes, within category & Poster beside an object & $0.69$--$0.72$ & $0.63$ \\
Checkpoint & Exclude monitor in the frame & $0.862$ & --- \\
Checkpoint and scenes & Exclude monitor in the frame & $0.802$ & --- \\
\bottomrule
\end{tabular}
\end{table}

A separate head on frozen DA-V2-Small features, excluding monitors in the frame, reaches
out-of-fold AUROC $0.863$. Retaining its most confident $70\%$ of inputs lowers failure
from $35.0\%$ to $18.7\%$.

The cross-view consistency \emph{baseline} runs the depth network on the right image as well.
This second network evaluation is specific to that baseline. It warps the right estimate into
the left frame with the reference disparity, aligns it like $H_D$, and scores its disagreement
with the left prediction inside the region. Within category, that disagreement predicts failure
at an AUROC of $0.62$ on sheets and $0.56$ on posters, and the spread of the twelve checkpoints'
aligned predictions at $0.62$ and $0.60$. Pooled values depend on the category mix. Without the monitor-in-the-frame category, both fall below $0.5$, at $0.378$ and $0.440$. With its rebuilt reference, the ensemble spread reaches $0.588$, and the cross-view check cannot be
computed on that pool. Both kinds of signal answer
when to distrust the prediction, and neither returns the surface to use instead.

\subsection{Scene-Level Statistics on Public Benchmarks}
\label{sec:supp-scene}

The pooled results of Sections~\ref{sec:res-motion} to~\ref{sec:res-adds} count region--checkpoint combinations. Table~\ref{tab:supp-scene} restates the results on public data per independent scene or photograph. Each interval gives the $2.5$th and $97.5$th percentiles of $5{,}000$ bootstrap resamples. A resample draws whole scenes or photographs with replacement, so all regions and checkpoints of one scene stay together. Under executed motion, the median surface error of the answer is below the network's in $42$ of the $44$ scenes and in every resample. Both categories improve: the median falls from $3.83\%$ to $2.55\%$ over the $23$ sheet scenes and from $5.14\%$ to $4.22\%$ over the $21$ poster scenes. On the stereo photographs, Intervene3D worsens fewer checkpoints than the detector on $361$ of the $404$ photographs and more on one. On ScanNet++ mirrors, the share of confidently wrong surfaces is lower in all $30$ scenes.

\begin{table}[htb]
\caption{\textbf{Scene-level statistics on public benchmarks.} Illusion denotes the illusion benchmark \citep{illusion2025}. Point estimates with bootstrap intervals over scenes or photographs. Surface made worse: share of combinations whose surface error exceeds the network's. Confidently wrong: Appendix~\ref{sec:supp-e8}. The detector is OWLv2 followed by the ring plane (Section~\ref{sec:res-adds}).}
\label{tab:supp-scene}
\centering
\footnotesize \setlength{\tabcolsep}{3pt}
\begin{tabular}{@{}lllcc@{}}
\toprule
Benchmark & Resampled unit & Quantity & \textcolor{baseink}{Network or detector} & \textcolor{oursink}{Intervene3D} \\
\midrule
Illusion, executed motion & $44$ scenes & median surface error & $4.35\%$ [$2.90$, $6.74$] & $\mathbf{2.98\%}$ [$1.98$, $5.32$] \\
Illusion, stereo & $404$ photographs & surface made worse & $52.6\%$ [$49.3$, $56.1$] & $\mathbf{0.8\%}$ [$0.4$, $1.3$] \\
ScanNet++ mirrors & $30$ scenes & confidently wrong & $35.3\%$ [$29.5$, $41.0$] & $\mathbf{5.0\%}$ [$3.3$, $7.0$] \\
\bottomrule
\end{tabular}
\end{table}

The relative reduction of the median surface error under executed motion is $31.5\%$, with a bootstrap interval of $16.5\%$ to $39.3\%$. The detector share of $52.6\%$ comes from the scorer that also produces Table~\ref{tab:supp-detectors}. The independent recompute cited in Section~\ref{sec:res-adds} reads the ring-plane error from a separate file and gives $52.7\%$.

\subsection{Mirror and Glass Detectors}
\label{sec:supp-detectors}

Section~\ref{sec:res-adds} scores an open-vocabulary detector on the surface that it returns. Table~\ref{tab:supp-detectors} adds a second open-vocabulary detector, Grounding DINO\footnote{Liu et al., Grounding DINO: Marrying DINO with grounded pre-training for open-set object detection, ECCV 2024.}, and two segmentation networks trained for mirrors and glass: HetNet\footnote{He, Lin, and Lau, Efficient mirror detection via multi-level heterogeneous learning, AAAI 2023.} with its two released weight sets, and GDNet \citep{glassdetection2020}. A segmentation network calls a region when it marks at least half of the region. Otherwise the panel call of OWLv2 applies. A called region is replaced by the ring plane. The five detector configurations make the network surface worse on $52.1$ to $53.6\%$ of the stereo combinations and on $24.3$ to $29.0\%$ under executed motion, against $0.8\%$ and $9.3\%$ for Intervene3D. The released weights operate as intended on real mirrors: on the $123$ frames of Mirror3D, HetNet with PMD weights marks $96.7\%$ of the mirrors, GDNet $87.8\%$, and HetNet with MSD weights $40.7\%$.

\begin{table}[htb]
\caption{\textbf{Detectors scored on the returned surface.} Share of region--checkpoint combinations whose surface error exceeds the network's, on the illusion benchmark: $4{,}848$ stereo combinations and $1{,}620$ combinations under executed motion. A flagged region is replaced by the ring plane.}
\label{tab:supp-detectors}
\centering
\footnotesize \setlength{\tabcolsep}{5pt}
\begin{tabular}{@{}lcc@{}}
\toprule
Method & Stereo & Executed motion \\
\midrule
OWLv2 \citep{owlv2} & $52.6\%$ & $25.8\%$ \\
Grounding DINO & $52.3\%$ & $27.8\%$ \\
HetNet (PMD weights), then OWLv2 & $52.1\%$ & $24.3\%$ \\
HetNet (MSD weights), then OWLv2 & $52.5\%$ & $25.5\%$ \\
GDNet, then OWLv2 & $53.6\%$ & $29.0\%$ \\
\midrule
\oursrow \oursname{Intervene3D} & $\mathbf{0.8\%}$ & $\mathbf{9.3\%}$ \\
\bottomrule
\end{tabular}
\end{table}

\subsection{Selection under Restricted Action Sets}
\label{sec:supp-sweep}

The hand-held mirror recording supplies $33$ candidate motions per region. With all of them available, maximin and summed selection choose the same motion on $318$ of the $324$ region--checkpoint combinations, because one motion separates every contending pair. The selection is therefore repeated with the action set restricted to a random subset of $k$ motions. For each $k$, $20$ subsets are drawn per region. Identifiability, selection, and the decision then use only the motions of the subset. The stored separabilities and residuals of every candidate motion are reused, and the full action set reproduces the counts of Section~\ref{sec:res-motion} exactly. Table~\ref{tab:supp-sweep} reports the share of correct decisions. Accuracy falls for both rules as $k$ shrinks, because fewer regions stay identifiable. Maximin is at least as accurate as summed selection in all $160$ combinations of subset size and draw: it is higher in $146$ and equal in $14$. The gap is largest at twelve motions, where it reaches $2.9$ percentage points. The gain appears where several hypothesis pairs contend, as the maximin criterion predicts. At twelve motions it reaches $14.8$ points for Metric3D-Large, where $27$ of $27$ regions have two or more contending pairs, $13.7$ points for Metric3D-Small ($26$ regions), $3.5$ for DA3-Large ($19$), $2.8$ for DA3-Small ($10$), and $0.2$ for DA3-Base ($4$). For the other seven checkpoints, at most two regions have more than one contending pair, and the two rules select the same motions.

\begin{table}[htb]
\caption{\textbf{Maximin and summed selection under restricted action sets.} Hand-held mirror recording, $324$ region--checkpoint combinations, $20$ random subsets of $k$ motions per region. Correct decisions are means over the subsets. The gap is given with its minimum and maximum over the $20$ draws.}
\label{tab:supp-sweep}
\centering
\footnotesize \setlength{\tabcolsep}{5pt}
\begin{tabular}{@{}cccc@{}}
\toprule
Motions $k$ & \oursname{Maximin} & Summed & Gap in points [min, max] \\
\midrule
$2$  & $48.3\%$ & $47.6\%$ & $+0.7$ [$0.0$, $+3.7$] \\
$3$  & $50.8\%$ & $49.5\%$ & $+1.3$ [$0.0$, $+4.9$] \\
$5$  & $56.4\%$ & $54.4\%$ & $+1.9$ [$+0.3$, $+4.3$] \\
$8$  & $60.1\%$ & $57.5\%$ & $+2.5$ [$+0.6$, $+3.7$] \\
$12$ & $63.0\%$ & $60.0\%$ & $\mathbf{+2.9}$ [$+1.2$, $+4.0$] \\
$16$ & $65.9\%$ & $63.5\%$ & $+2.4$ [$+0.9$, $+4.3$] \\
$24$ & $75.3\%$ & $73.5\%$ & $+1.7$ [$+0.3$, $+3.4$] \\
$33$ (all) & $87.7\%$ & $86.4\%$ & $+1.2$ \\
\bottomrule
\end{tabular}
\end{table}

\section{Robustness and Controls}
This appendix tests sensitivity to the separability threshold, the occlusion weight, the belief temperature, the commitment threshold, and the sensor scale. It then gives implementation details and the computational cost.

\subsection{Threshold Sensitivity}
\label{sec:supp-sensitivity}

The separability threshold is swept from $0.25$ to $3.0$~px on the three-mechanism benchmark, with the label and the decision recomputed at each value. Every reported quantity of Intervene3D-sum stays unchanged, at $56.9\%$ abstentions, a CEA of $0.431$, a committed accuracy of
$1.000$, and an FPCR of $0.000$. Only at $5.0$~px does the abstention rate move, to $57.4\%$.
Separability on this benchmark is bimodal, so the seven thresholds tested between $0.25$ and
$3.0$~px give identical aggregates. On real data the two thresholds were fixed before the runs
reported here, at $2.0$ gray levels for one stereo baseline and $1.0$~px for executed motions.

\subsection{Sensitivity to the Occlusion Weight}
\label{sec:supp-occlusion}

The composite distance of \eqref{eq:distance} weights an occlusion term against a motion term,
with $\lambda_o = 4.0$ throughout. The same eleven seeds are rerun at $\lambda_o \in \{1, 2, 4, 8, 16\}$, a $16\times$ range, with every other setting fixed. Table~\ref{tab:supp-occlusion}
reports the outcome for Intervene3D-sum. Committed accuracy stays at
$1.000$ at every weight, including on mirrors, and the weight moves only how many scenes are
answered. From $\lambda_o = 2$ upward the CEA stays between $0.430$ and $0.436$, and only at
$\lambda_o = 1$, where a single visibility change weighs no more than a pixel of motion, does
it fall to $0.379$.

\begin{table}[htb]
\caption{\textbf{Occlusion weight.} Intervene3D-sum at five occlusion weights $\lambda_o$, $11$ seeds, with the value used in the paper in bold. Largest-translation commits at $1.000$ and passive multi-view at $0.622$ at every weight.}
\label{tab:supp-occlusion}
\centering
\footnotesize \setlength{\tabcolsep}{4pt}
\begin{tabular}{@{}ccccc@{}}
\toprule
$\lambda_o$ & Committed & Reflection committed & CEA & Abstention \\
\midrule
$1$  & $1.000 \pm 0.000$ & $1.000$ & $0.379 \pm 0.051$ & $62.1\%$ \\
$2$  & $1.000 \pm 0.000$ & $1.000$ & $0.431 \pm 0.053$ & $56.9\%$ \\
\oursrow $\mathbf{4}$ & $\mathbf{1.000 \pm 0.000}$ & $\mathbf{1.000}$ & $\mathbf{0.431 \pm 0.062}$ & $\mathbf{56.9\%}$ \\
$8$  & $1.000 \pm 0.000$ & $1.000$ & $0.430 \pm 0.062$ & $57.0\%$ \\
$16$ & $1.000 \pm 0.000$ & $1.000$ & $0.436 \pm 0.063$ & $56.4\%$ \\
\bottomrule
\end{tabular}
\end{table}

\subsection{Threshold, Temperature and Sensor Scale}
\label{sec:supp-hyper}

Three constants stay fixed throughout and receive no tuning, namely the separability threshold
$\epsilon = 1.0$~px, the inverse temperature $\beta = 1.0$ of the belief update, and the
commitment threshold $\tau = 0.45$. A threshold stated in pixels depends on the sensor for its
meaning, and resolution is therefore swept alongside the other two constants.
Table~\ref{tab:supp-hyper} reports all three over the same eleven seeds. The constants $\beta$
and $\tau$ enter at evaluation, which lets those arms share one dataset under a paired
comparison. Resolution instead changes the rendered scene, and each resolution arm regenerates
its data, which also reseeds the draw, so those arms admit only an unpaired comparison.

\begin{table}[htb]
\caption{\textbf{Threshold, temperature and sensor scale.} Intervene3D-sum over eleven seeds, for the belief temperature $\beta$, the commitment threshold $\tau$, and sensor resolution. Committed accuracy is $1.000$ in every row. The resolution block admits an unpaired comparison alone.}
\label{tab:supp-hyper}
\centering
\footnotesize \setlength{\tabcolsep}{4pt}
\begin{tabular}{@{}lccc@{}}
\toprule
Configuration & CEA & Abstention & Passive multi-view \\
\midrule
\groupband{4}{Belief temperature $\beta$, paired}
$0.25$ & $0.369 \pm 0.055$ & $63.1\%$ & $0.622 \pm 0.030$ \\
$0.5$  & $0.430 \pm 0.061$ & $57.0\%$ & $0.622 \pm 0.030$ \\
\oursrow $\mathbf{1.0}$ \textbf{(reported)} & $\mathbf{0.431 \pm 0.062}$ & $\mathbf{56.9\%}$ & $0.622 \pm 0.030$ \\
$2.0$  & $0.431 \pm 0.062$ & $56.9\%$ & $0.622 \pm 0.030$ \\
$4.0$  & $0.431 \pm 0.062$ & $56.9\%$ & $0.622 \pm 0.030$ \\
\groupband{4}{Commitment threshold $\tau$, paired}
$0.30$ & $0.431 \pm 0.062$ & $56.9\%$ & $0.622 \pm 0.030$ \\
\oursrow $\mathbf{0.45}$ \textbf{(reported)} & $\mathbf{0.431 \pm 0.062}$ & $\mathbf{56.9\%}$ & $0.622 \pm 0.030$ \\
$0.60$ & $0.431 \pm 0.062$ & $56.9\%$ & $0.622 \pm 0.030$ \\
$0.80$ & $0.429 \pm 0.064$ & $57.1\%$ & $0.622 \pm 0.030$ \\
\groupband{4}{Sensor resolution and $\epsilon$, unpaired}
$160 \times 120$, $\epsilon = 0.5$~px & $0.471 \pm 0.063$ & $52.9\%$ & $0.648 \pm 0.034$ \\
$160 \times 120$, $\epsilon = 1.0$~px & $0.470 \pm 0.062$ & $53.0\%$ & $0.648 \pm 0.034$ \\
\oursrow $\mathbf{320 \times 240}$, $\boldsymbol{\epsilon = 1.0}$~\textbf{px (used)} & $\mathbf{0.463 \pm 0.063}$ & $\mathbf{53.7\%}$ & $0.626 \pm 0.051$ \\
$640 \times 480$, $\epsilon = 1.0$~px & $0.469 \pm 0.042$ & $53.1\%$ & $0.636 \pm 0.035$ \\
$640 \times 480$, $\epsilon = 2.0$~px & $0.469 \pm 0.042$ & $53.1\%$ & $0.636 \pm 0.035$ \\
\bottomrule
\end{tabular}
\end{table}

The CEA moves by $0.001$ across $\beta \geq 0.5$ and falls to $0.369$ only at $\beta = 0.25$,
where the discount on the evidence leaves more pairs in contention. Across $\tau$ it moves by
$0.002$. The identifiability rule of \eqref{eq:abstain} abstains first. With the uniform prior, the posterior after a separating action exceeds the tested values of $\tau$ in all but a
handful of cases, which are the only ones the commitment threshold governs. Across a fourfold
range of linear resolution the CEA moves by $0.008$, and holding $\epsilon$ in pixels or scaling
it with the resolution gives nearly identical outcomes, which follows from the bimodal
separability noted above.

\subsection{Implementation Details}
\label{sec:supp-impl}

\paragraph{Visibility and general distance.} A landmark is visible under a hypothesis
when it projects inside the image and the ray to it passes through the finite interface
aperture, and operator markers follow the same rule. A landmark counted as visible under one
mechanism and hidden under another therefore contributes to $D_{\text{occlusion}}$, and the term
stays discrete, so a single unambiguous appearance or disappearance carries the weight
described in Section~\ref{sec:separability}. The general implementation also supports
\[
D_{\mathrm{general}}=D+\lambda_gD_{\mathrm{geometry}}+\lambda_fD_{\mathrm{feature}},
\]
where $D_{\mathrm{geometry}}$ is root-mean-square 3D distance between back-projected landmarks
and $D_{\mathrm{feature}}$ is an appearance distance. Both weights are zero in every reported
experiment, giving the two-term distance in \eqref{eq:distance}.

\paragraph{Landmarks and belief.} Each synthetic scene carries $40$ content landmarks placed by
jittered stratified sampling over the inner $85\%$ of the interface extent, in a random subset
of $40$ cells of a $7 \times 7$ grid, together with $3$ operator markers. The camera measures
$320 \times 240$ with $f_x = f_y = 320$. Stratification ensures coverage of the region, and coverage determines whether a visibility change becomes observable at all. The belief update uses
$\beta = 1.0$ and a posterior floor of $10^{-6}$, which stops one unlucky observation from eliminating a hypothesis permanently.

\paragraph{Computational cost.} On the four-mechanism benchmark the selector enumerates
$|\mathcal{A}| = 33$ actions against six hypothesis pairs. Predicting every consequence in
closed form, with projection and visibility, and evaluating identifiability takes a median of
$46$~ms per scene in \textsc{NumPy} on one central processing unit core, without graphics
acceleration. On real photographs the depth forward pass dominates the cost, at $462$~ms per
image end to end for the smallest checkpoint and $1140$~ms for the largest. The cost grows
linearly in $|\mathcal{A}|$, and the evaluations remain independent, which allows a direct
parallelization over available cores.

\section{Scope, Limitations and Reproducibility}
This appendix discusses extensions beyond planar optics, the scope and limitations, reproducibility, and the use of generative tools.

\subsection{Extending Beyond Planar Optics}
\label{sec:supp-extension}

The four-member family stays planar in its interfaces, covering a flat panel, a planar mirror,
and a planar pane. Curved mirrors and non-planar transparent surfaces occur commonly, and the
present section states what changes to admit them and what remains.

\paragraph{Decision rule.}
No part of the decision rule depends on planarity. Separability \eqref{eq:separability} holds
between the predicted consequences of any two hypotheses under any action, identifiability
\eqref{eq:identifiability} maximizes that quantity over the action set, and the abstention rule
\eqref{eq:abstain} minimizes over contending pairs against a separability threshold. A curved
mirror enters as a further $H_k$ carrying its own transition $W$, and reaches the maximin
selector \eqref{eq:maximin} as one more pair without any change to the criterion. The belief
update \eqref{eq:belief} and the adequacy test behave the same way, since both require a
residual and nothing further. Appendix~\ref{sec:supp-e5} does exactly this with a spherical
mirror.

\paragraph{Transition guarantee.}
The present construction derives its value from closed-form transitions. A planar mirror
reflects exactly across a plane, and a plane-parallel pane forms a fixed apparent image in the
paraxial limit. Reference views therefore match to a largest deviation of order $10^{-13}$~px,
and any divergence after an action follows from the mechanism alone. A general curved surface
admits no exact transition, and two routes remain open, each weakening a different property.

The first route is a parametric surface, such as a spherical mirror of given sag. It keeps a closed-form transition and therefore keeps the matched guarantee. Such a family extends in a controlled way, although only to the surfaces that the parameterization covers and only with the right parameters. With the true sag it commits correctly on curved mirrors, and with a nominal sag it commits wrongly on $43\%$ of them (Appendix~\ref{sec:supp-e5}).

The second route is a learned transition. It covers arbitrary geometry but forfeits the property on which the design depends. Once a separate fit produces $W$ per mechanism, a difference between two predictions may follow from a difference between fitted models instead of physical mechanisms, which is exactly the confound that the shared projection removes. The matched-counterfactual guarantee would then hold only approximately, by an amount that is hard to bound.

The adequacy test addresses mechanisms that neither route models. A mechanism that no member explains leaves a residual that the test can flag. A mechanism that some member explains within the
threshold under every action stays indistinguishable from that member, which no test on the
observations can change.

\subsection{Scope and Limitations}
\label{sec:supp-limits}
\label{sec:supp-concurrent}

Five limits bound what the reported numbers establish. First, the synthetic identifiability
AUROC and the false physical certainty rate are contrasts rather than absolute scores. One
computation produces both the resolvability label and the decision on the synthetic benchmark,
so the AUROC of $1.000$ and the FPCR of $0.000$ in Table~\ref{tab:phase1} stand only against the
values that other methods reach on identical scenes. The noisy-encoder ablation, which breaks
that identity, keeps committed accuracy at $0.996$ and the FPCR at $0.000$. Second, identifiability
is relative to the family, the action set and $\epsilon$. A mechanism outside the family is
detected only when it leaves a residual above the adequacy threshold, which the water-like
surface of Appendix~\ref{sec:supp-e5} does on $4\%$ of scenes. Third, the illusion-benchmark motions come from frames captured for another purpose. On that benchmark the hypothesis set holds two members, so maximin and the summed objective coincide. The hand-held mirror recording supplies three contending hypotheses, and there maximin selection answers correctly on $87.7\%$, against $55.6\%$ for the largest motion and $33.3\%$ for a random motion. With the action set restricted, maximin is never less accurate than summed selection (Appendix~\ref{sec:supp-sweep}). Fourth, the evidence for the decision on real data comes from an illusion benchmark whose panels fool the predictors and from hand-held recordings of a monitor and a mirror. On mirrors without an operator cue the decision abstains, as identifiability predicts (Appendix~\ref{sec:supp-e8}). Fifth, the stereo
protocol fits the panel on a ring outside the region, which misplaces it where the panel stands in front of its surroundings, and alignment uses the reference outside the region at test time,
for the answer and the predictor alike.

Two contemporaneous works appeared during the preparation of this paper. \citet{mirage2026} probe hallucinated structure elicited by planar yet
perceptually ambiguous inputs, contributing a benchmark with context variation, two
second-order magnitude-based scores, and a training strategy that reduces the effect. The two
studies agree on the phenomenon and differ in the response offered, namely a reduction in
hallucination magnitude there and a decidability criterion with an explicit abstention here.
\citet{vggtworld2026} forecast the temporal evolution of frozen geometry-foundation features in place of future pixels. This result establishes that such features carry predictive structure and
admit direct operations. Appendix~\ref{sec:supp-learned} uses the same frozen substrate for a
different purpose, namely predicting whether a prediction fails.

\subsection{Reproducibility}
\label{sec:supp-repro}

All synthetic results are averaged over $11$ seeds with the benchmark redrawn per seed, and the sample standard deviation is reported throughout. Splits of the synthetic benchmark follow base
scenes, which keeps the variants of a scene inside one partition. The learned failure classifier is reported
both with the checkpoint alone held out and with the held-out checkpoint's scenes removed from
training as well (Appendix~\ref{sec:supp-learned}). The protocols on real data, including their thresholds, the
ring width, the admissibility rule and the acceptance rule of the rebuilt reference, were
written down before the runs they govern, and analyses added after a result are labeled as
such in the text. Every depth checkpoint is used exactly as released, and nothing is trained or tuned on any evaluation set. The output convention of each checkpoint is verified against reference stereo disparity before use, because the attribute conventionally named
\texttt{predicted\_depth} carries metric depth in some families and inverse depth in others, and
invalid pixels of a prediction stay invalid rather than becoming zero. One checkpoint,
UniDepth-v2, runs without its optimized attention kernel on the available hardware, and its
numbers form a lower bound on the released model. Code will be released upon acceptance.

\subsection{Use of Generative Artificial Intelligence}
\label{sec:supp-ai-disclosure}

We developed the research problem, conceptual framework, hypothesis family, methodological
architecture, theoretical results, and benchmark design independently, before using any
generative artificial intelligence tool. Such tools then served only as a general-purpose
assistant under our direct supervision, for implementation and routine programming, plotting
scripts, supplementary controls, dataset cleaning, and refinement of text that we had already
drafted. We reviewed every assisted output before inclusion, verified all citations against the
original sources, and obtained every reported number from experiments that we ran with the
released code.

\end{document}